\documentclass[lettersize,journal]{IEEEtran}
\usepackage{amsmath,amsfonts}
\usepackage{algorithmic}
\usepackage{algorithm}
\usepackage{array}
\usepackage[caption=false,font=normalsize,labelfont=sf,textfont=sf]{subfig}
\usepackage{textcomp}
\usepackage{stfloats}
\usepackage{url}
\usepackage{verbatim}
\usepackage{graphicx}
\usepackage{amssymb}
\usepackage{multirow}
\usepackage{booktabs}
\usepackage{colortbl}
\usepackage{color}
\usepackage{bbding}
\usepackage{balance} 
\usepackage{cite}
\begin{document}

\title{Towards Accurate Emotion-Attributed Video Captioning via

Fine-grained Emotion-Cause Pair Extraction}

\author{Weidong Chen,~\IEEEmembership{Member,~IEEE},
        Cheng Ye, 
        Zhendong Mao,~\IEEEmembership{Member,~IEEE},
        Liping Wang, \\
        Xinyan Liu,
        Yongdong Zhang,~\IEEEmembership{Fellow,~IEEE}
        
\thanks{This research is supported by Artificial Intelligence-National Science and Technology Major Project 2023ZD0121200, and the National Natural Science Foundation of China under Grants 62302474 and 62502115. Corresponding author: Zhendong Mao.}
\thanks{Weidong Chen, Cheng Ye, and Liping Wang are with the School of Information Science and Technology, University of Science and Technology of China, Hefei 230027, China (e-mail: chenweidong@ustc.edu.cn; kyrieye@mail.ustc.edu.cn; lipingwang@mail.ustc.edu.cn).}
\thanks{Zhendong Mao and Yongdong Zhang, are with the School of Information Science and Technology, University of Science and Technology of China, and are also with the Institute of Artificial Intelligence, Hefei Comprehensive National Science Center, Hefei 230027, China (e-mail: zdmao@ustc.edu.cn; zhyd73@ustc.edu.cn).}
\thanks{Xinyan Liu is with the School of Information Science and Technology, Harbin Institute of Technology (Weihai), Weihai 264209, China (e-mail: xinyliu@hit.edu.cn).}}

\markboth{Journal of \LaTeX\ Class Files,~Vol.~14, No.~8, August~2021}%
{Shell \MakeLowercase{\textit{et al.}}: A Sample Article Using IEEEtran.cls for IEEE Journals}


\maketitle

\begin{abstract}
Emotional Video Captioning (EVC) is a challenging task that aims to generate factually accurate and emotionally rich descriptions for videos. Existing EVC methods leverage holistic visual features to mine global emotional cues, and then aggregate multimodal features to guide the emotional caption generation, which ignores the critical characteristic of the EVC task. Visual emotions are evoked by specific motivational causes, which are usually only implied in core video segments. The holistic mining brings significant information redundancy and inaccurate emotional cues. Thus, fine-grained visual cause extraction has a facilitative effect on both emotion perception and emotion-attributed caption generation. To this end, we propose a fine-grained emotion-cause pair extraction framework for emotion-attributed video captioning. Specifically, we learn pair-wise emotion and cause features in two rounds: 1) We propose a Concept-aware Visual Semantic Decomposition module to augment visual features by exploring scene, object, and motion concepts. Besides, to enhance emotional features, we propose a Visual-guided Emotion Interpretable Learning module, which guides emotion refinement with visual temporal dynamics, and augments the interpretable refinement process by reliable VAD-vector constraints. 2) We achieve emotion-cause pair extraction by cross-coupling the visual and emotional features before and after refinement, and leverage contrastive loss to achieve semantic forced alignment. Overall, our approach optimizes complex semantic understanding and emotion perception of videos, leading to a promising performance in emotional captioning. Extensive experiments on three challenging datasets demonstrate the superiority of our approach and each proposed module, \emph{e.g.}, achieving the best performances with +4.4\% and +5.4\% w.r.t. BLEU-2 and ROUGE-L, respectively, on the EVC-MSVD dataset.
\end{abstract}


\section{Introduction}
\IEEEPARstart{W}{ith} the rapid evolution of self-media platforms, users are no longer satisfied with mere information transmission, they are increasingly inclined to express subjective emotions through multimodal contents. This burgeoning demand for emotional expression has catalyzed the community’s enthusiasm for exploring automated visual emotion analysis, \emph{i.e.,} facial emotion recognition\cite{fard2025affectnet+,lan2025expllm,guo2024benchmarking,chen2026subjective}, video emotion classification \cite{wang2025mdkat,li2025feature,qi2024versatile,ye2025multi,wang2025combatting}, and emotional video captioning (EVC) \cite{liu2025enriched, yu2024prompting,yu2023comprehensive, deng2021syntax}. Among them, Emotional Video Captioning (EVC) task, which aims to generate factually accurate and emotionally rich descriptions for video, stands out as a fundamental yet highly challenging endeavor due to the complex intersection of vision, emotion, and language studies. Firstly, EVC necessitates the effective alignment and fusion of content across multiple modalities (e.g., vision and text). Besides, it requires to not only comprehend factual content but also accurately perceive implicit emotional cues within the video and seamlessly integrate them into caption generation. EVC offers significant potential for various real-world applications, \emph{i.e.,} empathetic conversational agent, video comment generation, and automated film production.

\begin{figure}[t]        
\center{\includegraphics[width=0.9\linewidth] {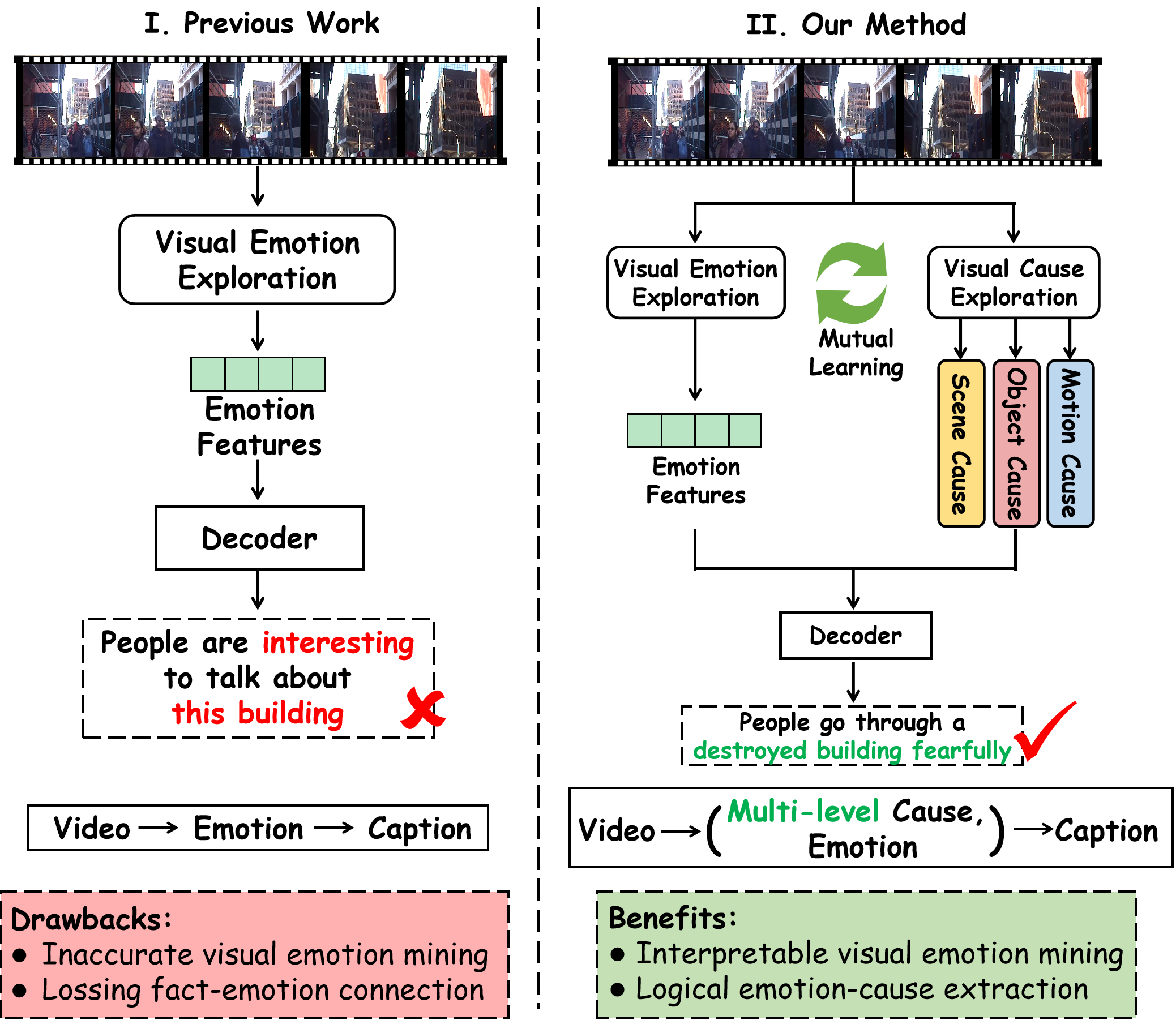}}
\caption{Motivation of our paper, which illustrates the difference between previous work and our proposed method, and shows two classical failure cases of previous work.}
\label{fig1}
\vspace{-15pt}
\end{figure}

Existing EVC methods \cite{wang2021emotion, song2022contextual, song2023emotion,fu2024sentiment,gao2025learning,ye2025improving,ye2024dual,xu2024rule,xu2024cross,huang2025graph} usually leverage holistic visual features to mine global emotional cues, and then aggregate multimodal features to guide the emotional caption generation. However, they ignore the critical characteristic of the EVC task. Visual emotions are evoked by specific motivational causes, which are usually only implied in core video segments. The holistic mining brings significant information redundancy and emotional bias. As shown in Fig. \ref{fig1}, previous works overlook the critical contribution of visual causes that the building is destroyed, resulting in wrong emotional descriptions `interesting'. In fact, exploring the visual causes not only enhances the accuracy and interpretability of emotion perception, but also facilitates the factual and emotional description generation, alleviating the mutual interference of multi-source information during emotional caption generation. Thus, in this paper, we focus on multimodal emotion-cause pair extraction to enhance the EVC task. The key differences between the previous works and ours are illustrated in Fig. \ref{fig1}.

Current research on multimodal Emotion-Cause Pair Extraction (ECPE) remains predominantly centered on linguistic cause cues, encountering significant challenges in processing visual causes. Although recent studies attempt to integrate visual or auditory features, they primarily focus on text-based extraction and treat other modalities merely as auxiliary information, failing to achieve modality-agnostic cause-emotion alignment. Conversely, our approach focuses on the pairwise extraction of visual-based causes(e.g., scenes, objects, actions) and textual emotional expressions within multimodal scenarios, which transcends the limitations of existing methods and is more challenging. Specifically, it brings two inherent difficulties: (1) \textbf{Cause-to-Emotion Mismatch} — Visual information is frequently characterized by high redundancy and contains significant noise that is irrelevant to emotional cues. Such coarse-grained visual representations fail to provide the necessary support for fine-grained emotional mining. (2) \textbf{Emotion-to-Cause Asynchrony} — subtle emotion changes are hard to spatio-temporal synchrony with their visual causes in videos. Thus, the difficulty of our work lies in how to extract fine-grained cause features and mine subtle emotional cues for collaborative emotion-cause pair extraction. 

To fill the research gap, we propose a Fine-grained Emotion-Cause Pair Extraction framework (MM-ECPE++) in this paper for joint extraction of emotional cues and visual causes through multi-round step-by-step refinement by mutual learning for emotion-attributed video captioning. Specifically, in the first-round mutual learning, we propose a Concept-aware Visual Semantic Decomposition (CVSD) and a Visual-guided Emotion Interpretable Learning (VEIL) to achieve preliminary refinement on visual features and emotion lexicon to eliminate the noise caused by emotion-irrelevant spatio-temporal visual information and video-irrelevant emotional information.

Subsequently, we capture the visual emotion cues and causes in the second-round mutual learning. Specifically, we couple the preliminary refined features and original features by a cross-attention mechanism. Besides, we design a contrastive loss to force semantic alignment in pairwise extraction. Overall, our method optimizes the complex semantic and emotional understanding of videos, leading to a promising performance in emotional captioning.

\begin{figure*}[t]        
\center{\includegraphics[width=0.98\linewidth] {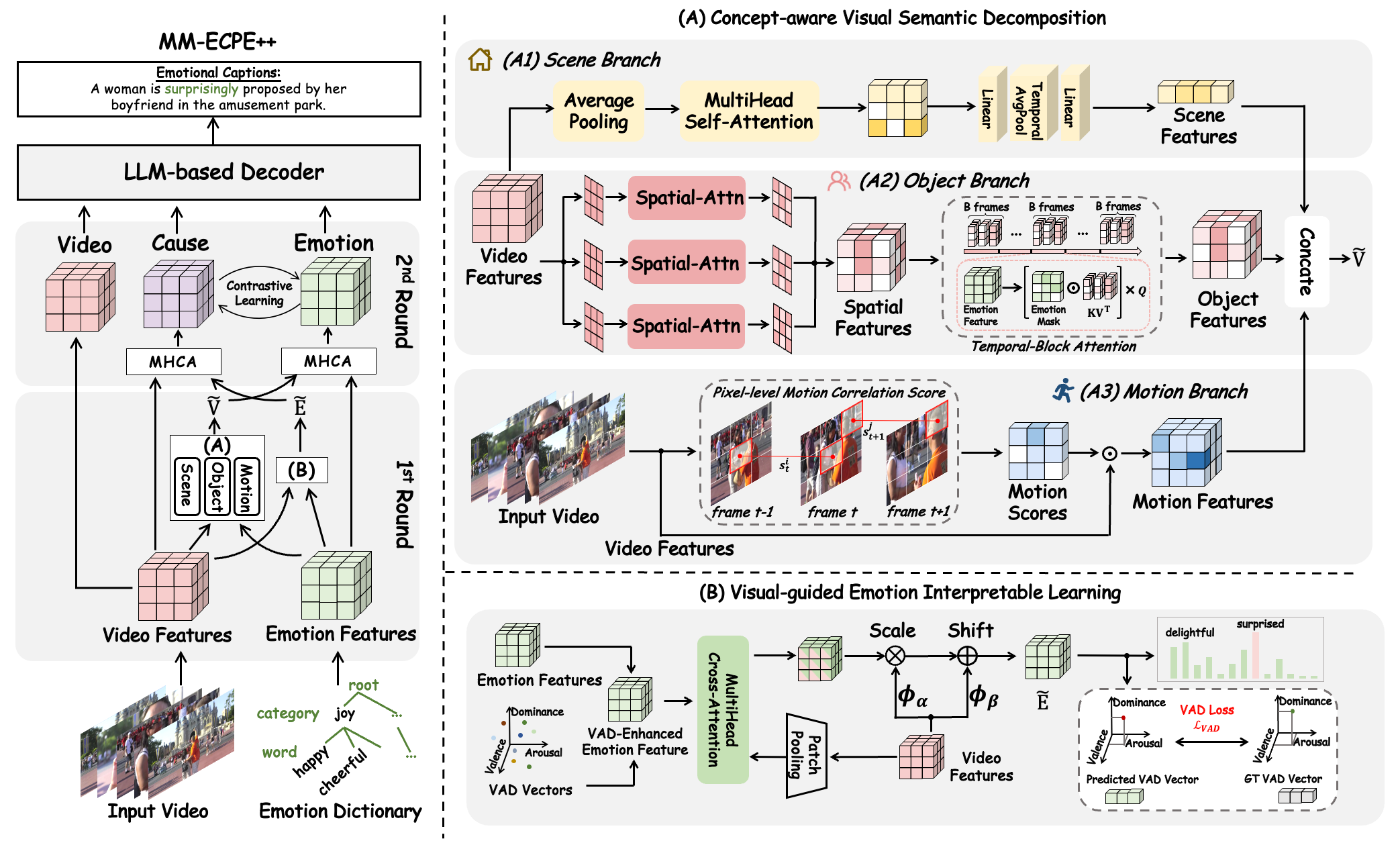}}
\vspace{-10pt}
\caption{Overview of our proposed MM-ECPE++ framework for emotion-attributed video captioning. Given an input video and emotion dictionary, we perform concept-level fine-grained visual cause-emotion pair extraction by two modules: (A) Concept-aware Visual Semantic Decomposition (top right plot) and (B) Video-guided Emotion Interpretable Learning (bottom right plot). Subsequently, we perform emotion-cause pair mutual refinement by two cross-attention components and a contrastive loss. Finally, pairwise emotion-cause features are combined with video features to be sent into the emotion-attributed caption generation decoder.}
\vspace{-5pt}
\label{fig2}
\end{figure*}

In short, our main contributions are summarized as follows:

$\bullet$ In this paper, we propose a novel Fine-grained Emotion-Cause Pair Extraction network (MM-ECPE++) for emotion-attributed video captioning, which focuses on noticing the visual causes of emotions, improving the accuracy and interpretability of emotion perception and effectively alleviating the mutual interference of multiple factors in captioning.

$\bullet$ To achieve cross-modal pair extraction (visual causes + textual emotions), we propose a novel multi-round mutual learning framework. It preliminarily refines the relevant information in the video and emotion lexicon in the first round, and then, the refined information interacts with each other's original features, and is jointly optimized by contrastive loss to force the alignment of the common semantic space, to capture the emotion-cause pair.

$\bullet$ Technically, we propose a Concept-aware Visual Semantic Decomposition (CVSD) and a Visual-guided Emotion Interpretable Augmentation (VEIL) to achieve preliminary refinement on video features and emotion lexicon to eliminate the noise caused by emotion-irrelevant spatio-temporal visual information and video-irrelevant emotional information. 

$\bullet$ Experiments on three public datasets, \emph{e.g.}, EVC-MSVD, EVC-VE, and EVC-Combined, illustrate the superiority of our method, \emph{e.g.}, achieving the best performance with +4.4\% and +5.4\% w.r.t. BLEU-2 and ROUGE-L, respectively, on EVC-MSVD.

This manuscript extends MM-ECPE\cite{ye2025multi}, which was selected as an Oral paper at ACM Multimedia 2025. While sharing the same core motivation, we extend this work in the following aspects:
1) We completely rebuild the framework MM-ECPE into MM-ECPE++, which significantly improves the mutual learning through finer-grained factual refinement and more interpretable emotional exploration.
2) We improve the factual refinement by redesigning a Concept-aware Visual Adaptive Refinement module. Specifically, we enhance the granularity of visual refinement to the concept level, extracting key cause cues from three categories of concepts: scenes, objects, and motions.
3) We improve the emotional exploration by redesigning an Interpretable Emotion Exploration module. Specifically, we introduce an emotional VAD dictionary and design a multi-dimensional constrained emotional loss to improve the interpretable emotional exploration.
4) We extend the Related Work section by adding more analyses about the new research on emotion-cause pair extraction and emotional video captioning, and making more discussions about the differences between them and our work.
5) We perform the main experiment and ablation study on a larger EVC-Combine dataset instead of the original EVC-MSVD dataset, which helps to validate the robustness and generalization ability of our method.
6) We add more ablation studies to quantitatively validate the effectiveness of our proposed modules, as shown in Tables \ref{component}-\ref{nce}.
7) We discuss the effects of the trade-off parameters in Fig. \ref{figloss} to evaluate the sensitivity of our model to parameters.
8) We improve the qualitative results to illustrate the effectiveness of our proposed modules, as shown in Fig. \ref{fig3}-\ref{fig5}.
9) We add human evaluation to assess the degree of human preference of our generated descriptions, which is difficult to reflect in objective metrics such as BLEU.

\section{Related Work}
\subsection{Emotion-cause Pair Extraction}
Emotion cause analysis has received increasing attention due to its potential to promote emotional exploration. Emotion-cause pair extraction(ECPE) is one of the most representative tasks \cite{tang2021graph,li2022ecpec, wang2024enhancing, wang2024enhancing, an2023global,song2025towards,lin2024prompting,10756281}. Xia and Ding \cite{xia2019emotion} firstly propose this task to extract emotions and their causes in documents. To this day, recent research has focused more on analyzing emotion-cause pairs in diverse textual scenarios such as articles, stories, microblogs, or conversations. Zhu \emph{et al.} \cite{zhu2024knowledge} propose a novel knowledge-guided graph attention network, which designs two guiding principles: inter-clause dependency graph and inter-pair possibility graph, to aggregate features between clause pairs for emotion-clause and cause-clause pair extraction. Chen \emph{et al.} \cite{chen2024coarse} propose to utilize the causal discourse knowledge in a knowledge distillation way, which designs a teacher model to learn to predict causal connective words between utterances and then guide a student model in identifying both the fine-grained emotion labels and causal spans. 
Hu \emph{et al.} \cite{hu2023emotion} analyze that emotion extraction is more crucial to the ECPE task than cause extraction and propose an approach oriented toward emotion prediction, which aims to fully exploit the potential of emotion prediction to enhance ECPE. Furthermore, they design a synchronization mechanism to share their optimizations in the training process.

Besides, some researchers have extended the ECPE task to the multimodal field \cite{li2024multimodal,jin2024improving,lin2024prompting,jin2024improving,liu2024bootstrapping, wang2022multimodal, ma2024extraction, ju2025enhanced,huang2025graph,jin2024improving,jin2024improving2,gao2023vectorized}. 
Wang \emph{et al.} \cite{wang2024observe} design a two-stage framework, which first generates emotion-cause aware video captions and then facilitates the generation of emotion causes. With the help of corresponding captions, the model improves the accuracy of pair extractions. However, all existing multimodal ECPE still perform on the text modality while treating image and audio modalities as auxiliary information. In contrast, our work focuses on the cross-modal scenario (visual causes + textual emotions).

\subsection{Emotional Video Captioning}
Emotional video captioning task is proposed to solve the problem of boring and soulless captions generated by traditional video captioning model, which firstly mines emotional clues hidden in videos and collaborates visual contents to generate more vivid captions. Wang \emph{et al.} \cite{wang2021emotion} firstly propose this task and release a new dataset. They design two independent prediction networks for factual and emotional captioning separately and finally generate emotional descriptions by weighted average scores. Song \emph{et al.} \cite{song2022contextual} propose a contextual attention network to recognize and describe the fact and emotion in the video by semantic-rich context learning. Song \emph{et al.} \cite{song2023emotion} propose a novel tree-structured emotion learning module to achieve explicit emotion perception. Song \emph{et al.} \cite{song2024emotional} incorporate visual context, textual context, and visual-textual relevance into an aggregated multimodal contextual vector to enhance video captioning. Ye \emph{et al.} \cite{ye2024dual} propose a dual-path collaborative generation network to dynamically perceive emotions at different time steps and design an emotion-adaptive decoder to solve the problem of overemphasis on the role of emotional guidance. Wang \emph{et al.} \cite{wang2025emotion} introduce two learnable prompting strategies: visual emotion prompting and textual emotion prompting, to learn emotional cue representations and further design two levels of objective functions: the ER-sentence level and the AU-word level alignment losses, to facilitate the interaction and alignment. 

All aforementioned works directly perceive the emotional cues from visual contents and ignore that emotional cues have intrinsic motivational causes reflected in the video content. To the best of our knowledge, we are the first work to notice the visual causes of emotions, improving the accuracy and interpretability of emotion perception and effectively alleviating the mutual interference of multiple factors in emotional caption generation.
\section{Method}

In this section, we describe our proposed Fine-grained Emotion-Cause Pair Extraction framework (MM-ECPE++) for the EVC task. As shown in Fig. \ref{fig2}, it mainly consists of five parts: Multimodal Feature Extraction, Concept-aware Visual Semantic Decomposition, Visual-guided Emotion Interpretable Learning, Emotion-Cause Pair Extraction, and Emotion-Attributed Caption Generation.

\subsection{Multimodal Feature Extraction}\label{3.1}
{\noindent\textbf{Video Encoder.} 
Following the setting of previous works \cite{song2023emotion, ye2024dual}, we firstly down-sample the video to obtain a frame sequence $\{f_1,f_2…,f_T\}$, where $T$ denotes the number of frames. Subsequently, we leverage the pre-trained Transformer-based feature encoder \cite{yu-etal-2024-eliciting} to extract visual features:}

\begin{align}
    \mathcal{V}&= {\rm VisEncoder([f_1,f_2,…,f_T])\in\mathbb{R}^{T \times P \times D}},
\end{align}
{where ${\rm VideoEncoder}$ denotes the visual feature projector and $D$ is the feature dimension. $P$ is the number of patches since each frame is divided into $P$ patches.}

{\noindent\textbf{Emotion Encoder.} Following previous works \cite{song2023emotion, ye2024dual}, we leverage a pre-defined psychology emotional vocabulary dictionary $\Omega=\{w_i\}_{i=1}^{N_w}$, where $w_i$ denotes the $i$-th emotion word, such as ``happy'', and $N_w$ denotes the number of emotion words in dictionary. Then, we leverage the pre-trained Transformer-based emotion embedding, \emph{i.e.,} GloVe \cite{pennington2014glove}, to encode these emotion words to ${\mathcal{E}}\in \mathbb{R}^{N_w\times D}$. The whole process can be formalized as:}
\begin{equation}
    \mathcal{E} = {\rm EmoEncoder}(\Omega)\in \mathbb{R}^{N_w\times D}.\\
\end{equation}
{${\mathcal{E}}$ contains rich emotion candidates and helps to mine accurate emotional cues from video contents.}

{Subsequently, in the first-round mutual learning, we propose two novel modules, \emph{i.e.,} Concept-aware Visual Semantic Decomposition (CVSD) and Video-guided Emotion Interpretable Learning (VEIL), to perform preliminary refinement and filter irrelevant information on video features and emotion lexicon, respectively, which provides strong factual and emotional support for emotion-cause pair extraction.}

\subsection{Concept-aware Visual Semantic Decomposition}\label{3.2}

For emotion-cause pair extraction, the video contains a lot of redundant information and only the key semantic components are effective. Thus, it is a considerable challenge to efficiently capture key semantic information in video representations. Conventional methods often employ frame sampling techniques\cite{wang2023improving,chen2022multi,song2023emotion,song2022contextual} or design a frame selection module\cite{ye2024dual,ye2025multi,chen2021cascade,chen2023weakly} to eliminate redundant information. However, these methods essentially only perform filtering and reweighting at the frame-level, and struggle to truly decouple semantic entities in the visual space. Therefore, we propose a novel Concept-aware Visual Semantic Decomposition module to decouple distinct semantic entity types from the video for better emotion-cause pair extraction. Specifically, we distinguish three types of key semantic entities: (1) \textbf{Scene feature} includes background information and reflects the environment of the video. (2) \textbf{Object feature} focuses on the main static elements (\emph{i.e.,} characters and things) relevant to the emotional event in the video. (3) \textbf{Motion feature} focuses on dynamic elements that can evoke emotions (\emph{i.e.,} the movements and trajectories of objects). With these semantic entities, we could eliminate the redundant information in the video and accurately capture highly interpretable emotional causes.

{\textbf{Scene Feature.} Scene features primarily focus on the environment and background context within a video, playing a crucial role in predicting coarse-grained emotion color. For example, a rainy environment typically reflects negative emotions (\emph{i.e.,} sadness or fear), which lays the foundation for subsequent fine-grained emotion prediction. Meanwhile, we observe that capturing scene features tends to focus more on global information rather than local details. Therefore, we first perform an average pooling operation along the patch dimension to obtain global video features:}
\begin{equation}
    \mathcal{V}_{agg} = {\rm AvgPool(V)}\in \mathbb{R}^{T\times D},
\end{equation}
{where ${\rm AvgPool}$ denotes the average pooling operation. Subsequently, we perform a multi-head self-attention between frames to capture global context dependencies:}
\begin{equation}
    \widetilde{\mathcal{V}}_{agg} = {\rm MHSA}(\mathcal{V}_{agg}),
\end{equation}
{where ${\rm MHSA}$ denotes the multi-head self-attention operation. Finally, we capture scene features from global visual information by linear transformations and average pooling operations:}
\begin{equation}
    \mathcal{F}_{scene} = \phi_{scene}^2({\rm AvgPool}(\phi_{scene}^1(\widetilde{\mathcal{V}}_{agg})))\in\mathbb{R}^D,
\end{equation}
{where $\phi_{scene}^1, \phi_{scene}^2$ are two linear transformation layers and ${\rm AvgPool}$ denotes the average pooling operation, respectively.}

{\textbf{Object Feature.} Object features mainly capture static elements in a video (\emph{i.e.,} characters and things), which are the main carriers of emotional events. With accurate object features, our model could answer who/what evokes emotions, especially in hard samples with complex character relationships and multiple emotion types. Specifically, we refine visual features from spatial and temporal dimensions to capture object features. Firstly, we employ a spatial-geometry attention mechanism for each frame, which captures key object regions from horizontal and vertical spatial directions:}
\begin{align}
v_i = \mathcal{V}[i]|_{T-axis}&\in\mathbb{R}^{P\times D}\\
    Q_i^{axis} = W^{axis}_Qv_i, K_i^{axis} &= W^{axis}_Kv_i, V_i = W_Vv_i,\\
    O_i^{axis} = {\rm Softmax}(Q_i^{axis}&(K_i^{axis})^T), axis\in\{x,y\},\\
    O_i = O_i^y(O_i^xV_i)^T&\in\mathbb{R}^{P\times D},\\
    \mathcal{V}_{s} = [O_1,O_2,\cdots,&O_T]\in\mathbb{R}^{T\times P\times D},
\end{align}
{where $Q_i^{axis}(K_i^{axis})^T, axis\in\{x,y\}$ means calculate attention scores along vertical and horizontal axis for the $i$-th visual frame and $[,]$ denotes the concatenate operation. The spatial-geometry attention captures the key regions in each frame. However, not all frames are closely related to emotional arousal. Therefore, we next perform emotion-related frame filtering in the temporal dimension. Specifically, we generate an emotion mask $A$ by the emotion dictionary $\mathcal{E}$ to filter emotion-related frames. Meanwhile, to alleviate the long-range dependency of distant frames and reduce computational complexity, we leverage the temporal-block attention and apply the emotion mask to filter emotion-related frames. Specifically, each B frames is divided into a block, and a mask attention mechanism is executed within the block. Finally, the attention outputs of all blocks are concatenated to restore the shape:}
\begin{align}
    \mathcal{V}_{b,i} = \overline{\mathcal{V}_s}[iB:(i+1)B],i\in\{1,\cdots,T/B\},\\
    O_{b,i} = {\rm MaskAttn}(\mathcal{V}_{b,i},{\rm mask}=A)|_{\{Q,K,V\}:\mathcal{V}_{b,i}},\\
    \mathcal{F}_{object} = [O_{b,1},O_{b,2},\cdots,O_{b,T/B}]\in\mathbb{R}^{T\times P\times D},
\end{align}
{where $\mathcal{V}_{b,i}$ denotes the visual features of the $i$-th block and ${\rm MaskAttn}$ denotes the mask attention mechanism. Finally, we obtain the object features $\mathcal{F}_{object}$ with detailed spatial region awareness and accurate emotion-related temporal filtering, which play a crucial role in analyzing emotional affiliation.}

{\textbf{Motion Feature.} Due to the variable and dynamic nature of video modality, it is insufficient for mining emotional causes to capture static backgrounds and elements solely through learning scene and object features. Therefore, we propose to learn the action features to capture temporal dynamics of moving elements (\emph{i.e.,} actions and activities). Specifically, we propose a pixel-level motion correlation calculation module, which quantifies motion information through calculating the similarity of all pixel positions between adjacent frames. For the $j$-th patch of the $t$-th frame $v_t^j$, the motion information is quantified as:}
\begin{align}
    v_t^j &= \mathcal{V}[t]|_{T-axis}[j]|_{P-axis}\in\mathbb{R}^D,\\
    s_t^j &= 1-\frac{v_{t-1}^{j}\cdot v_t^j}{||v_{t-1}^{j}||||v_t^{j}||},
\end{align}
{where $s_t^j\in [0,2]$ denotes the degree of motion change. The lower the similarity of corresponding patches between adjacent frames, the greater the degree of motion change. By performing the calculation for each patch, we obtain the motion-aware mask matrix for $t$-th frame, denoted as \textbf{s}$_t\in\mathbb{R}^{P}$. Meanwhile, in order to eliminate biases caused by some unexpected disturbances (\emph{i.e.,}camera shake or perspective change), we calculate the global motion baseline through the average of all frames. The motion information for each frame comprehensively considers the global motion baseline and its own motion mask matrix. Finally, we leverage the motion information to generate motion features:}
\begin{align}
    &\overline{s} = \frac{1}{T}\sum_{i=1}^{T}s_t,\\
    &m_t = (1-\lambda_m)s_t+\lambda_m \overline{s},\\
    &\mathcal{F}_{motion} = m\odot{\mathcal{V}}\in\mathbb{R}^{T\times P \times D},
\end{align}
{where $\lambda_m$ is a hyper-parameter to control the weight of the global motion baseline. }

Finally, we obtain the refined visual features by aggregating three concept semantics:
\begin{align}
    &\mathcal{C} = [\mathcal{F}_{scene},\mathcal{F}_{object},\mathcal{F}_{motion}],\\
    \tilde{\mathcal{V}} = &\phi_{agg}^2({\rm AvgPool}(\phi_{agg}^1(\mathcal{C})))\in\mathbb{R}^{T\times P\times D},
\end{align}
{where $\phi_{agg}^1, \phi_{agg}^2$ denote two linear transformation layers and $[,], {\rm AvgPool}$ denote the concatenate operation and average pooling operation, respectively.}

\subsection{Visual-guided Emotion Interpretable Learning}\label{3.3}
We adopt the VAD lexicon\cite{mohammad2018obtaining} to obtain a 3-dimensional affective vector for each emotion word, i.e., $\mathbf{r}=(v,a,d)^{\top}\in\mathbb{R}^{3}$, where each entry is normalized to $[-1,1]$ (and out-of-lexicon words are set to $\mathbf{0}$). 
These psychologically grounded VAD priors provide an interpretable attribute space; thus, we leverage them to augment emotion representations and to enforce explicit alignment in the $(v,a,d)$ dimensions, improving interpretability of emotion learning.

Building on this, our visual-guided emotion interpretable learning consists of three components: VAD Modeling, Visual-guided Emotion Refinement, and VAD Regression Objective.

\noindent\textbf{VAD Modeling. } Given the emotion vocabulary $\Omega=\{w_i\}_{i=1}^{N_w}$, the NRC VAD lexicon provides a VAD triplet $(v_i,a_i,d_i)$ for each emotion word. We stack these triplets into a VAD feature matrix:

\begin{align}
\mathbf{F}^{vad}
&= \mathrm{VADEnc}(\{(v_i,a_i,d_i)\}_{i=1}^{N_w})
\in \mathbb{R}^{N_w\times 3}.
\end{align}

We then fuse $\mathbf{F}^{vad}$ with the emotion representations $\mathcal{E}\in\mathbb{R}^{N_w\times D}$ via concatenation, followed by a linear projection to integrate affective priors into the emotion space:
\begin{align}
\hat{\mathcal{E}}
&= W^{E}(\mathcal{E} \oplus \mathbf{F}^{vad}) + \mathbf{b}^{E}
\in \mathbb{R}^{N_w\times D},
\end{align}
where $\oplus$ denotes concatenation and $(W^{E}, \mathbf{b}^{E})$ are learnable parameters. 
The resulting $\hat{\mathcal{E}}$ serves as a \emph{VAD-enhanced emotion feature}, which augments semantic emotion features with explicit affective attributes.

\noindent\textbf{Visual-guided Emotion Refinement.} To refine emotion representations with both affective priors and visual evidence, we leverage video features to calibrate the VAD-enhanced emotion features. Specifically, we apply the compact video features $\mathcal{V}_{agg}$ on $\hat{\mathcal{E}}$ by a cross-attention mechanism:

\begin{align}
\mathcal{E}'
&= \mathrm{Attn}(\hat{\mathcal{E}},\,\mathcal{V}_{agg})
|_{Q:\mathcal{V}_{agg},\,\{K,V\}:\hat{\mathcal{E}}}
\in \mathbb{R}^{T\times D}.
\end{align}

To further incorporate global video context, we adopt a video-conditioned affine transformation to calibrate the refined emotion features. Specifically, scaling and shifting factors are generated from the video features via pooling operations:
\begin{align}
\boldsymbol{\alpha}
&= \mathrm{Pooling}|_{\mathrm{PD\text{-}axis}}
(\mathrm{Reshape}[\phi_{\alpha}(\mathcal{V})]|_{\mathbb{R}^{T\times P\times D}})
\in \mathbb{R}^{T\times 1}, \\
\boldsymbol{\beta}
&= \mathrm{Pooling}|_{\mathrm{P\text{-}axis}}
[\phi_{\beta}(\mathcal{V})]
\in \mathbb{R}^{T\times D}.
\end{align}

The final interpretable emotion representation is obtained as:
\begin{align}
\tilde{\mathcal{E}}
&= \boldsymbol{\alpha} \odot \mathrm{Norm}(\mathcal{E}') + \boldsymbol{\beta},
\end{align}
where $\mathrm{Norm}(\cdot)$ denotes a normalization operation applied to stabilize feature scale before affine modulation, and $\odot$ represents element-wise multiplication.

\noindent\textbf{VAD Regression Objective. }
To explicitly regularize the refined emotion representations with VAD attributes, we introduce a VAD regression head implemented as a single linear layer. We predict a VAD vector as follows:
\begin{align}
\hat{\mathbf r} = W^{V}\tilde{\mathcal{E}} + \mathbf b^{V},
\end{align}
where $\hat{\mathbf r}=(\hat v,\hat a,\hat d)^{\top}$ denotes the predicted VAD vector, and $W^{V}$ and $\mathbf b^{V}$ are learnable weights and bias.
We adopt mean squared error (MSE) to minimize the discrepancy between predicted VAD vectors and ground-truth VAD vectors:
\begin{align}
\mathcal{L}_{vad} = \left\|\hat{\mathbf r}-\mathbf r\right\|_{2}^{2},
\end{align}
where $\mathbf r=(v,a,d)^{\top}$ denotes the ground-truth VAD vector and $\left\|\cdot\right\|_{2}^{2}$ denotes the 2-norm operation.

\subsection{Emotion-Cause Pair Extraction}\label{3.4}
In the second round of mutual learning, in order to collaboratively extract emotion-cause pairs, we propose to leverage the refined visual features and the refined emotion features, which record the effective information of their modalities. With the interaction of the original features, the refined features are further refined to better represent visual causes and textual emotions. Thus, specifically, we exploit the cross-attention of the preliminary refined features and the original features to obtain the ultimate emotional cues $\mathbb{E}$ and visual causes $\mathbb{C}$. The formulations are: 
\begin{align}
    \mathbb{E} &= {\rm Attn}(\widetilde{\mathcal{V}},{\mathcal{E}})|_{Q:\widetilde{\mathcal{V}},\{K,V\}:{\mathcal{E}}}\in \mathbb{R}^{T\times P\times D},\\
    \mathbb{C} &= {\rm Attn}(\mathcal{V},\widetilde{\mathcal{E}})|_{Q:\widetilde{\mathcal{E}},\{K,V\}:{\mathcal{V}}}\in \mathbb{R}^{T\times P\times D},
\end{align}
where $\mathbb{E}$ and $\mathbb{C}$ are perceived by visual features and emotional features respectively, which ensures that they promote each other through collaborative extraction, thereby enhancing the accuracy and interpretability of emotion perception and helping decouple factual and emotional description generation.

Besides, to forcibly align the semantic representation of $[\mathbb{E}, \mathbb{C}]$ pairs, driven by the success of contrastive learning in visual-language pre-training tasks\cite{radford2021learning, fu2024linguistic}, we design an optimization objective functions based on intrinsic relationships between each emotion-cause pair. Specifically, given a batch of $B$ emotion-cause pairs, 
we maximize the cosine similarity of the emotion and cause embedding from $B$ real pairs in the batch while minimizing the cosine similarity of embeddings from the ($B^2 - B$) incorrect pairs:
\begin{equation}\begin{aligned}
  \mathcal{L}_{c2e}&=-\frac{1}{B}\sum_{i \in B}\log\frac{{\rm exp}{\{\mathbb{C}_{i}\cdot\mathbb{E}_{i}^{\top}/\tau\}}}{\sum_{j \in B}{\rm exp}{\{\mathbb{C}_{i}\cdot\mathbb{E}_{j}^{\top}/\tau\}}}, \\
  \mathcal{L}_{e2c}&=-\frac{1}{B}\sum_{i \in B}\log\frac{{\rm exp}{\left(\mathbb{E}_{i}\cdot\mathbb{C}_{i}^{\top}/\tau\right)}}{\sum_{j \in B}{\rm exp}{\left(\mathbb{E}_{i}\cdot\mathbb{C}_{j}^{\top}/\tau\right)}}, \\
 & \mathcal{L}_{ctr}=\frac{1}{2}(\mathcal{L}_{c2e}+\mathcal{L}_{e2c}),
\end{aligned}\end{equation}
where $\tau$ is the temperature hyper-parameter.

\subsection{Emotion-Attributed Caption Generation}\label{3.5}
After obtaining sufficient semantic information including video and emotion-cause pair features, we concatenate them to generate a unified multi-modal semantic representation $M=[\mathcal{V};\mathbb{C};\mathbb{E}]\in\mathbb{R}^{(T\times 3P)\times D}$. Subsequently, we utilize a pre-trained BLIP-2 \cite{yu-etal-2024-eliciting} architecture, which contains Q-former and pre-trained language decoder \cite{raffel2020exploring}, to generate captions. Specifically, we leverage a Q-former module to convert $M$ to tokens that could be interpreted by a language decoder. The Q-former uses learnable query vectors $q\in \mathbb{R}^{N_q\times D}$ to extract interpretable representations, where $N_q$ is the query length. We first use self-attention to aggregate each query:
\begin{equation}
    \overline{q}={\rm Attn}(q)|_{\{Q,K,V\}:q}+q\in \mathbb{R}^{N_q\times D},
\end{equation}
then the aggregated query is applied to generate interpretable multimodal features by a cross-attention and a linear projection:
\begin{equation}
        \overline{M} = \phi_M[{\rm Attn}(\overline{q},M)|_{Q:\overline{q},\{K,V\}:M}+\overline{q}]\in \mathbb{R}^{N_q\times D},
\end{equation}
Finally, we send $\overline{M}$ to the language decoder for caption generation.

In addition to $\mathcal{L}_{ctr}$, we also design two emotion-aware optimization objectives. Firstly, since the important role of emotional words in EVC, we use emotion-focused cross-entropy loss that adds a penalty term on emotional words:
\begin{equation}
    \mathcal{L}_{e}=\begin{cases}-(1+\theta)\sum_{t=1}^{N_T}\log P(y_{t}|y_{<t}),&{\rm if}\ y_{t}\in \Omega,\\\\-\sum_{t=1}^{N_T}\log P(y_{t}|y_{<t}),&{\rm otherwise},\end{cases}
\end{equation}
where $\theta$ is a hyper-parameter that controls the level of punishment in the equation when $y_t$ is emotional words from emotional lexicon $\Omega$, and $N_T$ denote the sequence length.

Furthermore, to provide more sufficient emotional loss, we also design an emotional classification loss. Specifically, we add a simple classification header for the emotion features $E$ to obtain the emotional distribution $d$. Then we calculate the emotional classification loss as follows:
\begin{equation}
    \mathcal{L}_{cls}=-\sum_{e \in \Omega}\log P(e|d),
\end{equation}
where $e$ is emotional words from emotional lexicon $\Omega$. The overall loss is the weighted sum of four losses:

\begin{equation}
\mathcal{L}
=\lambda_{e}\mathcal{L}_{e}
+\lambda_{cls}\mathcal{L}_{cls}
+\lambda_{ctr}\mathcal{L}_{ctr}
+\lambda_{vad}\mathcal{L}_{vad},
\end{equation}
where $\lambda_{e}$, $\lambda_{cls}$, $\lambda_{ctr}$, and $\lambda_{vad}$ are hyper-parameters that balance the four objectives. 
Our model is trained end-to-end by minimizing the overall loss $\mathcal{L}$.

\begin{table*}[t]
\centering
\caption{\label{main}{Performance for emotional video captioning task on three benchmarks. The best results are highlighted in bold.}}
\scalebox{0.95}{
\begin{tabular}{l|cc|ccccccc|cc}
\toprule
\multirow{2}{*}{\textbf{Methods}}  &  \multicolumn{2}{c|}{\textbf{Emotion}} &\multicolumn{7}{c|}{\textbf{Semantic}} & \multicolumn{2}{c}{\textbf{Hybrid}}\\
~ &${\rm Acc}_{sw}$$\uparrow$& ${\rm Acc}_c$$\uparrow$&{BLEU-1$\uparrow$}   & {BLEU-2$\uparrow$}   & {BLEU-3$\uparrow$}   & {BLEU-4$\uparrow$}   & {METEOR$\uparrow$} & {ROUGE$\uparrow$} & {CIDEr$\uparrow$}&BFS$\uparrow$&CFS$\uparrow$  \\ 
\toprule
\multicolumn{12}{c}{\textbf{EVC-MSVD}}\\
\cmidrule{1-12}
{SA-LSTM} \cite{wang2018reconstruction}  &68.8 &67.2 &80.7 &67.9 &56.3 &45.5 &33.0 &68.2 &72.1 &59.0 &71.3\\
FT        \cite{wang2021emotion}        &69.4 &67.1 &77.2 &60.3 &47.4 &36.3 &29.0 &63.4 &62.5 &52.5 &63.7\\
SGN       \cite{ryu2021semantic}        &73.9 &73.1 &77.5 &62.7 &51.3 &41.1 &30.6 &63.6 &71.0 &56.4 &71.5\\
CANet     \cite{song2022contextual}     &78.7 &76.8 &78.5 &64.0 &52.1 &41.8 &30.8 &65.7 &74.4 &57.9 &75.1\\
{VEIN}    \cite{song2024emotional}     &82.7 &82.1 &82.0 &68.4 &57.1 &45.9 &33.0 &69.0 &79.6 &62.4 &80.2\\
{EPAN}    \cite{song2023emotion}       &84.1 &82.8 &82.5 &69.6 &57.8 &46.2 &34.4 &69.8 &80.6 &63.1 &81.1\\
{DCGN}    \cite{ye2024dual}            &${86.5}$  &${85.7}$& ${84.5}$ & ${70.9}$ & ${59.2}$ & ${48.7}$ & ${35.7}$&${71.0 }$   & ${85.2}$  & ${65.7}$&${86.6}$ \\
{MM-ECPE}\cite{ye2025multi}&${{90.4}}$ &{89.1} & {96.9}	&{87.8}	&{80.2}	&{71.4}	&{45.5}	&{83.1}	&${168.3}$ &${81.8}$	&${152.6}$ \\
\textbf{Ours}&\textbf{91.2} &\textbf{89.7} &\textbf {97.7}	&\textbf{88.4}	&\textbf{80.5}	&\textbf{71.5}	&\textbf{45.8}	&\textbf{83.8}	&\textbf{169.6} &\textbf{82.3}	&\textbf{153.8} \\
\toprule
\multicolumn{12}{c}{\textbf{EVC-VE}}\\
\cmidrule{1-12}
{SA-LSTM} \cite{wang2018reconstruction} & 	48.6 &47.1& 71.0 &51.1 &34.5 &22.5 & 19.6&40.7 &30.2& 38.9& 33.7 \\
CANet \cite{song2022contextual} &41.9 &39.7 &66.9 &44.8 &29.3 &19.3 &18.2 &37.9 &23.3 &33.9 &26.8\\
{VEIN} \cite{song2024emotional}  & 57.4 &56.8&71.6&52.1&37.4&26.3&20.9&41.7&33.4&43.0& 39.2\\
{EPAN} \cite{song2023emotion}  & 63.8 &62.3& 73.6& 54.0 &38.3 &27.0 & 21.2&42.3 &34.7&45.0 &40.4\\
{DCGN} \cite{ye2024dual}  & 	${71.0}$  &${69.4}$& ${74.5}$ & ${55.3}$ & ${40.0}$ & ${28.1}$ & ${23.4}$&${47.7}$    &${41.5}$& ${47.3}$&${46.9}$  \\	
{MM-ECPE}\cite{ye2025multi} & {73.4}	&{72.3}& {76.8}	&{57.5}	&{41.7}	&{28.9}	&{24.7}	&{49.5}	&{65.2}	 &{49.2}	&{66.7}	  \\
\textbf{Ours}&{\textbf{74.1}} &\textbf{72.9} & \textbf{77.8}	&\textbf{58.2}	&\textbf{42.2}	&\textbf{29.2}	&\textbf{25.1}	&\textbf{50.4}	&\textbf{66.1} &\textbf{49.7}	&\textbf{67.6} \\
\toprule
\multicolumn{12}{c}{\textbf{EVC-Combine}}\\
\cmidrule{1-12}
{SA-LSTM} \cite{wang2018reconstruction} &	53.4 &50.7& 70.6& 51.4& 36.7& 25.4 & 21.0&45.9& 38.8& 41.2 &41.5 \\
FT \cite{wang2021emotion}&51.2 &49.6& 67.6 &47.2 &32.0 &21.6 &20.4 &43.1 &29.0& 37.6 &33.3\\
SGN\cite{ryu2021semantic}&50.4&48.6&68.7&48.9&34.2&24.0&20.1&44.8&35.5&39.1&38.3\\
CANet \cite{song2022contextual} &53.7 &52.7 &68.1 &47.7& 32.9 &22.5 &19.7 &43.7 &34.5 &38.8 &38.2\\
{VEIN} \cite{song2024emotional}& 59.0 &57.6&72.1&52.8&37.9&27.1&21.6&46.8&39.4&43.6& 43.1\\
{EPAN} \cite{song2023emotion} & 	69.3 &67.2& 74.4 &55.6& 39.9 &28.0 & 23.0&47.1& 43.0&47.0 &48.0\\
{DCGN} \cite{ye2024dual}  & 	${74.8}$ &${73.1}$ & ${75.6}$& ${56.7}$ & ${40.5}$ & ${28.5}$ & ${24.9}$&${51.7}$    & ${49.8}$& ${48.5}$ &${51.7}$  \\
{MM-ECPE}\cite{ye2025multi} & {75.6}	&{73.8}&{78.1}&{58.5}&{42.3}	&{30.2}	&{26.4}	&{53.8}	&{67.9}	&{50.4}	&{69.3} \\
\textbf{Ours} &{\textbf{76.4}} &\textbf{74.3} & \textbf{78.9}	&\textbf{59.2}	&\textbf{42.6}	&\textbf{30.5}	&\textbf{26.8}	&\textbf{54.5}	&\textbf{69.0} &\textbf{50.8}	&\textbf{70.3} \\
\bottomrule
\end{tabular}}
\end{table*}

\begin{table*}[t]
\centering
\caption{\label{component}{{The results of ablation studies on the EVC-Combine dataset to discuss the effectiveness of different inputs for the decoder.}}} 
\scalebox{0.85}{
\begin{tabular}{l|cc|ccccccc|cc}
\toprule
{Inputs}  &  ${\rm Acc}_{sw}$$\uparrow$& ${\rm Acc}_c$$\uparrow$&{BLEU-1$\uparrow$}   &{BLEU-2$\uparrow$}   & {BLEU-3$\uparrow$}   & {BLEU-4$\uparrow$}   & {METEOR$\uparrow$} & {ROUGE$\uparrow$} & {CIDEr$\uparrow$} &BFS$\uparrow$ &CFS$\uparrow$  \\
\midrule
Video  & 70.4& 68.9& 75.1& 56.0& 40.8& 29.2& 25.3& 52.0& {65.6}& 48.0&{66.4} \\
 Video+Emotion   & 74.6	&	72.8 &  76.4	& 56.9 	& 41.5  &	29.6  &	 25.9	& 53.1	& 66.6	& 49.4	& 68.0  \\
 Cause+Emotion & 75.4	&73.3	 &  77.2	& 57.7 	& 41.8  &	30.0  &	 26.2	& 53.7	& 68.1	& 49.9	& 69.4  \\
 Video+Cause+Emotion $\quad$  &{\textbf{76.4}} &\textbf{74.3} & \textbf{78.9}	&\textbf{59.2}	&\textbf{42.6}	&\textbf{30.5}	&\textbf{26.8}	&\textbf{54.5}	&\textbf{69.0} &\textbf{50.8}	&\textbf{70.3} \\
\bottomrule
\end{tabular}}
\end{table*}

\section{Experiment}
\subsection{Experimental Settings}
\textbf{Datasets. }We experiment on three public emotional video captioning benchmarks, \emph{e.g.}, EVC-MSVD\cite{chen2011collecting}, EVC-VE\cite{jiang2014predicting}, and EVC-Combine\cite{wang2021emotion}. \textbf{EVC-MSVD} contains 240/134 videos and 8,169/4,611 sentences for training/testing by additionally embedding emotion words into the caption sentences of the traditional video captioning dataset MSVD \cite{chen2011collecting}. \textbf{EVC-VE} is built based on a traditional emotional video prediction dataset VideoEmotion-8 \cite{jiang2014predicting} and is divided into 1,141/382 videos and 19,398/6,527 sentences for training/testing, respectively. \textbf{EVC-Combine} is the combination of EVC-MSVD and EVC-VE, which contains 1,381/516 videos and 27,567/11,138 sentences for training/testing.

\textbf{Evaluation Metrics. }To evaluate the accuracy of emotion in generated sentences effectively, following previous work\cite{wang2021emotion}, we consider the emotion word accuracy ${\rm Acc_{sw}}$ \cite{wang2021emotion} and emotion sentence accuracy ${\rm Acc_{c}}$ \cite{wang2021emotion}. Additionally, to measure the semantic matching degree between the generated sentence and the ground-truth label, we use the standard metrics, \emph{e.g.}, \textbf{BLEU}\cite{papineni2002bleu}, \textbf{METEOR}\cite{banerjee2005meteor}, \textbf{ROUGE}\cite{lin2004rouge}, and \textbf{CIDEr}\cite{vedantam2015cider}. Moreover, there are two hybrid metrics \textbf{BFS}\cite{wang2021emotion} and \textbf{CFS}\cite{wang2021emotion} that combine the emotion evaluation with BLEU and CIDEr metrics, respectively.

\textbf{Implementation Details.} For each video, we sample $T=20$ frames and resize them to $224\times 224$ with central cropping for pixel values. We extract video features by pre-trained VideoBLIP\cite{yu-etal-2024-eliciting}, and set the patch number to $P=16\times16$. Following \cite{wang2021emotion} and \cite{song2023emotion}, we build an overall vocabulary of size 32,128 that contains all words in the corpus and set the number of emotion words to $N_w=179$. The embedding dimensions are constructed to $d_E=300$ and $D=1408$. The block number is set to $B=5$. For the Q-former component, we set the length of the learnable queries to 32. In training, we adopt freeze FLAN-T5 XL \cite{raffel2020exploring} as the language decoder and integrate a LoRA adapter \cite{hu2021loralowrankadaptationlarge} with $r=16$ and $\alpha=32$ to align the visual and emotional latent space. We adopt the Adam optimizer \cite{kingma2014adam} with a learning rate of 7e-4, and the batch size is set to 16. We set the global motion parameter $\lambda_m$, the penalty parameter $\theta$, the objective function weights $\lambda_e$, $\lambda_{cls}$, $\lambda_{ctr}$, and $\lambda_{vad}$ to 0.2, 0.1, 0.8, 0.3, 0.2, and 0.6. The maximum length of captions is set to 15 and the size of beam search is set to 4. All experiments are implemented on 2 NVIDIA A800 GPUs.

\textbf{Baselines.} We consider several state-of-the-art methods to make comparisons and divide them into two categories: (i) Traditional Captioning Methods and (ii) Emotional Captioning Methods.

\begin{table*}[t]
\centering
\caption{\label{module}{{The results of ablation studies on EVC-Combine dataset to discuss the effectiveness of our proposed components, where CVSD, VEIL denote Concept-aware Visual Semantic Decomposition and Visual-guided Emotion Interpretable Learning, respectively.}}}
\scalebox{0.85}{
\begin{tabular}{cc|cc|ccccccc|cc}
\toprule
\multicolumn{2}{c|}{Components}  &  \multirow{2}*{${\rm Acc}_{sw}$$\uparrow$}& \multirow{2}*{${\rm Acc}_c\uparrow$}&\multirow{2}*{BLEU-1$\uparrow$} &\multirow{2}*{BLEU-2$\uparrow$}&\multirow{2}*{BLEU-3$\uparrow$}&\multirow{2}*{BLEU-4$\uparrow$}     & \multirow{2}*{METEOR$\uparrow$}   &\multirow{2}*{ROUGE$\uparrow$}& \multirow{2}*{CIDEr$\uparrow$}&\multirow{2}*{BFS$\uparrow$}&\multirow{2}*{CFS$\uparrow$}  \\
CVSD &VEIL &  &  &    & & & & & & & & \\ 
\midrule
{$\times$}&{$\times$}& 70.4& 68.9& 75.1& 56.0& 40.8& 29.2& 25.3& 52.0& {65.6}& 48.0&{66.4} \\
{$\checkmark$}&{$\times$}& 71.5  & 69.8   & 77.6    & 58.0   & 42.1   & 30.2   & 26.4   &  53.9  &  68.4  &  49.4  & 68.9  \\
{$\times$}&{$\checkmark$}& 74.7  &  73.1  &  76.1  &  56.6  &  41.0  &  29.3  &  25.7  & 52.8   &  66.2  &  49.1  & 67.7  \\
{$\checkmark$}&{$\checkmark$}    &{\textbf{76.4}} &\textbf{74.3} & \textbf{78.9}	&\textbf{59.2}	&\textbf{42.6}	&\textbf{30.5}	&\textbf{26.8}	&\textbf{54.5}	&\textbf{69.0} &\textbf{50.8}	&\textbf{70.3} \\
\bottomrule
\end{tabular}}
\end{table*}

\begin{table*}[t]
\centering
\caption{\label{dense}{{The results of ablation studies on EVC-Combine dataset to discuss the effectiveness of each dense component in SOM Aware Visual Semantic Decomposition module.}}}
\scalebox{0.85}{
\begin{tabular}{ccc|cc|ccccccc|cc}
\toprule
\multicolumn{3}{c|}{Components}  &  \multirow{2}*{${\rm Acc}_{sw}$$\uparrow$}& \multirow{2}*{${\rm Acc}_c\uparrow$}&\multirow{2}*{BLEU-1$\uparrow$} &\multirow{2}*{BLEU-2$\uparrow$}&\multirow{2}*{BLEU-3$\uparrow$}&\multirow{2}*{BLEU-4$\uparrow$}     & \multirow{2}*{METEOR$\uparrow$}   &\multirow{2}*{ROUGE$\uparrow$}& \multirow{2}*{CIDEr$\uparrow$}&\multirow{2}*{BFS$\uparrow$}&\multirow{2}*{CFS$\uparrow$}  \\
Scene &Object&Motion &  &  &    & & & & & & & & \\ 
\midrule
{$\times$}&{$\times$}&{$\times$}& 74.7  &  73.1  &  76.1  &  56.6  &  41.0  &  29.3  &  25.7  & 52.8   &  66.2  &  49.1  & 67.7  \\
{$\checkmark$}&{$\times$}&{$\times$}& 75.2   & 73.4   &  75.6  &  55.9  &  40.1  &  28.4  & 24.9   &  52.2  &  65.8  &  48.6  & 67.5  \\
{$\times$}&{$\checkmark$}&{$\times$}& 74.4  & 73.0   & 77.2   &  57.5  &  41.5  &  29.5  & 25.9   &  53.2  &  66.6  &  49.5  &  68.0 \\
{$\times$}&{$\times$}&{$\checkmark$}& 74.1  & 72.8   &  76.5  &  57.0  &  41.2  &  29.3  & 25.9   &  53.4  & 67.0   &  49.2  & 68.3  \\
{$\times$}&{$\checkmark$}&{$\checkmark$}& 74.3  &  73.0  &  78.3  &  58.6  &  42.1  &  30.0  & 26.3   & 53.9   & 68.2   &  50.1  & 69.3  \\
{$\checkmark$}&{$\times$}&{$\checkmark$}& 75.5  &  73.8  &  77.0  &  57.2  &  41.4  &  29.5  & 25.7   &  53.0  &  66.4  & 49.6   & 68.1  \\
{$\checkmark$}&{$\checkmark$}&{$\times$}& 75.9  & 74.0   &  77.8  &  58.3  &  41.8  &  29.8  & 26.1   &  53.6  &  67.9  &  50.1  &  69.3 \\
{$\checkmark$}&{$\checkmark$}&{$\checkmark$}    &{\textbf{76.4}} &\textbf{74.3} & \textbf{78.9}	&\textbf{59.2}	&\textbf{42.6}	&\textbf{30.5}	&\textbf{26.8}	&\textbf{54.5}	&\textbf{69.0} &\textbf{50.8}	&\textbf{70.3} \\
\bottomrule
\end{tabular}}
\end{table*}

\smallskip
\noindent\textit{(i) Traditional Captioning Methods}

1) \textbf{SA-LSTM} (CVPR18') \cite{wang2018reconstruction} proposes a reconstruction network to leverage both the forward (video to sentence) and backward (sentence to video) flows for video captioning. The forward flow produces the sentence description based on the encoded video semantic features and the backward flow reproduces the video features based on the hidden state sequence generated by the decoder.

2) \textbf{CANet} (TMM22') \cite{song2022contextual} proposes a contextual attention network to recognize and describe the fact and emotion by semantic-rich context learning, which first extracts visual and textual features from both video and previously generated words and then applies the attention mechanism to capture informative contexts for captioning.

3) \textbf{VideoBLIP} (EMNLP24') \cite{yu2024eliciting} proposes a training paradigm that induces in-context learning over video and text by capturing key properties of pre-training data found by prior work to be crucial for improving the ability of LLMs to generate video descriptions.

\smallskip
\noindent\textit{(ii) Emotional Captioning Methods}

1) \textbf{FT} (TMM21') \cite{wang2021emotion} introduces a visual emotion analyzer to detect implicit emotional cues and a dual-stream network that integrates emotional and factual semantics for caption generation by a weighted sum operation.

2) \textbf{EPAN} (MM23') \cite{song2023emotion} introduces a tree-structured emotion repository to enable hierarchical emotion perception and an emotion-prior awareness network to achieve the explicit and fine-grained emotion perception by an emotion masking mechanism and then decode the emotional caption by exploiting the multimodal semantic cues.

3) \textbf{VEIN} (TIP24') \cite{song2024emotional} designs a vision-based emotion interpretation network, which first models the emotion distribution over an open psychological vocabulary and then incorporates visual context, textual context, and visual-textual relevance into an aggregated multimodal contextual vector to enhance video captioning.

4) \textbf{DCGN} (MM24') \cite{ye2024dual} propose a framework to leverage fact-emotion collaborative learning to address the challenge of dynamic emotion changes during caption generation. 

5) \textbf{MM-ECPE} (MM25')\cite{ye2025multi} focus on the importance of emotional
causes in emotional exploration and propose a multi-round mutual learning network to jointly extract emotion-cause pairs for LLM-based caption generation.

\subsection{Comparison with State-of-the-art Methods}
As shown in Table \ref{main}, we present the comparison with the state-of-the-art methods on three EVC datasets. We evaluate the performance of our model and have the following meticulous observations:

\begin{table*}[t]
\centering
\caption{\label{nce}{The results of ablation studies on EVC-Combine dataset to discuss the effectiveness of $\mathcal{L}_{ctr}$ and $\mathcal{L}_{vad}$ on different settings. 1st, 2nd denote applying contrastive loss $\mathcal{L}_{ctr}$ on the 1st round mutual learning ($\widetilde{\mathcal{V}},\ \widetilde{\mathcal{E}}$) or 2nd round mutual learning ($\mathbb{C}$, $\mathbb{E}$), respectively.}} 
\scalebox{0.85}{
\begin{tabular}{cc|c|cc|ccccccc|cc}
\toprule
\multicolumn{2}{c|}{Components}& \multirow{2}*{Settings}& \multirow{2}*{${\rm Acc}_{sw}$$\uparrow$}& \multirow{2}*{${\rm Acc}_c$$\uparrow$}&\multirow{2}*{BLEU-1$\uparrow$}   & \multirow{2}*{BLEU-2$\uparrow$}   & \multirow{2}*{BLEU-3$\uparrow$}   & \multirow{2}*{BLEU-4$\uparrow$}   & \multirow{2}*{METEOR$\uparrow$} & \multirow{2}*{ROUGE$\uparrow$} & \multirow{2}*{CIDEr$\uparrow$}&\multirow{2}*{BFS$\uparrow$}&\multirow{2}*{CFS$\uparrow$}  \\
$\mathcal{L}_{ctr}$ &$\mathcal{L}_{vad}$& &  &  &    & & & & & & & & \\ 
\midrule
{$\times$}&{$\times$}&-&  73.6 &  71.9  &  76.8  &  57.6  &  41.1  &  28.9  &  25.4  &  52.1  & 65.3   &  49.0  &  66.8 \\
{$\times$}&{$\checkmark$}&-& 75.1  &  73.2  &  77.2  &  57.9  &  41.4  &  29.1  &  25.6  &  52.5  &  65.9  &  49.5  &  67.6 \\
\cmidrule{1-14} 
{$\checkmark$}&{$\checkmark$}&1st& 74.8  & 73.0   &  78.0  &  58.6  &  41.9  &  29.6  &  25.8  &  52.9  &  66.4  &  49.9  & 67.9  \\
{$\checkmark$}&{$\checkmark$}&1st+2nd& 75.7  & 73.8   &  78.5  &  58.9  &  42.2  &  30.0  &  26.3  &  53.6  &  67.8  & 50.4   &  69.2 \\
{$\checkmark$}&{$\checkmark$}& 2nd &{\textbf{76.4}} &\textbf{74.3} & \textbf{78.9}	&\textbf{59.2}	&\textbf{42.6}	&\textbf{30.5}	&\textbf{26.8}	&\textbf{54.5}	&\textbf{69.0} &\textbf{50.8}	&\textbf{70.3} \\
\bottomrule
\end{tabular}}
\end{table*}

\begin{table}[t]
\centering
\caption{\label{clip}{Comparison with the state-of-the-art methods with CLIP Score metric on EVC-MSVD dataset. The best results are highlighted in bold. $+\Delta\%$ is the improvement compared with our method.}}
\resizebox{0.9\columnwidth}{!}
{
\begin{tabular}{l|cl|cl}
\toprule
{Methods}  & {CLIPScores$\uparrow$}&{$+\Delta\%$$\uparrow$}&{RefCLIP-S$\uparrow$}&{$+\Delta\%$$\uparrow$}\\  
\midrule
VEIN \cite{song2024emotional} &59.8 &-   &81.9   & - \\
EPAN \cite{song2023emotion}   &61.1 &+2.2\% &83.2   & +1.6\% \\
DCGN \cite{ye2024dual}        &66.4 &+11.0\% &86.1   &+5.1\%  \\
MM-ECPE\cite{ye2025multi} &69.1 &+15.6\%  &88.6 &+8.2\%  \\
\textbf{Ours}&\textbf{69.7} &\textbf{+16.6\%} &\textbf{89.5}   & \textbf{+9.3\%} \\
\bottomrule
\end{tabular}}
\end{table}

Firstly, we test our model on semantic metrics, which could reflect the accuracy and vividness of generated captions. We observe that our model outperforms all existing methods on most semantic metrics. For instance, on the EVC-Combine dataset, our model performs better than DCGN\cite{ye2024dual} by 4.4\%/5.4\% on BLEU-2/ROUGE-L metrics, respectively. Especially on the CIDEr metric, our model achieves a huge improvement of 38.6\%. Due to the fine-grained visual concept refinement, our model could mine accurate factual semantics and provide customized guidance for caption generation, which generates more semantically relevant descriptions.

Secondly, our model also achieves the best performance on emotion accuracy, which is an extremely important metric to evaluate whether our model could predict accurate emotions. For example, our model performs better than MM-ECPE\cite{ye2025multi} by +1.0\%/+0.8\% with ${\rm Acc}_{sw}$/${\rm Acc}_{c}$ on the EVC-VE dataset. The proposed model utilizes visual temporal dynamics to guide emotion refinement and augments the interpretable refinement process by reliable VAD-vector constraints, which enhances the emotional understanding capabilities and predicts emotions more accurately.

Moreover, we evaluate our model on hybrid metrics, which investigate the overall performance of our model. As shown in the last two columns of Table \ref{main}, our model achieves the best performance with BFS and CFS metrics on three datasets, \emph{i.e., }+1.5\%/+1.4\% improvements over MM-ECPE\cite{ye2025multi} on EVC-VE/EVC-Combine datasets with CFS metric, which demonstrates that captions generated by our model take into account both emotional and factual accuracy.

In addition, we evaluate our model with two higher human correlation evaluation metrics: CLIPScore and RefCLIP-S, which focus on the degree of semantic alignment between the video/reference sentence and the prediction sentence. As shown in Table \ref{clip}, our model also achieves the best performance on these two metrics, \emph{i.e.,} +5.0\%/+3.9\% improvements over DCGN\cite{ye2024dual} with CLIPScore/RefCLIP-S metric on EVC-MSVD dataset, which demonstrates that captions generated by our model are not only similar to the reference, but also highly semantic-aligned with the video and reference.

\subsection{Ablation Studies}
\noindent\textbf{Discussion on emotion-cause pairs. }To explore the impact of emotion-cause pairs on model performance, we conduct an ablation study on different inputs for caption generation, \emph{i.e.,} video, cause, and emotion features. As shown in Table \ref{component}, we firstly observe that compared with only sending video features into decoder, adding the emotion cue features could significantly enhance the emotional perception ability, \emph{i.e.,} +6.0\%/+5.7\% improvements on ${\rm Acc}_{sw}$/${\rm Acc}_{c}$. The pre-trained language decoder lacks the ability to understand the emotional cues implicit in video. Instead, our module firstly locates the emotion-related visual features by concept-aware visual refinement and leverages them to mine more accurate visual emotion cues. Besides, adding the cause features could significantly enhance the semantic understanding, \emph{i.e.,} +3.3\%/+3.6\% improvements on BLEU-1/CIDEr metrics. Our model achieves accurate descriptions of key visual elements by precisely mining fine-grained cause concepts, thereby improving the performance of semantic metrics.

\begin{figure*}[t]        
\center{\includegraphics[width=0.95\linewidth] {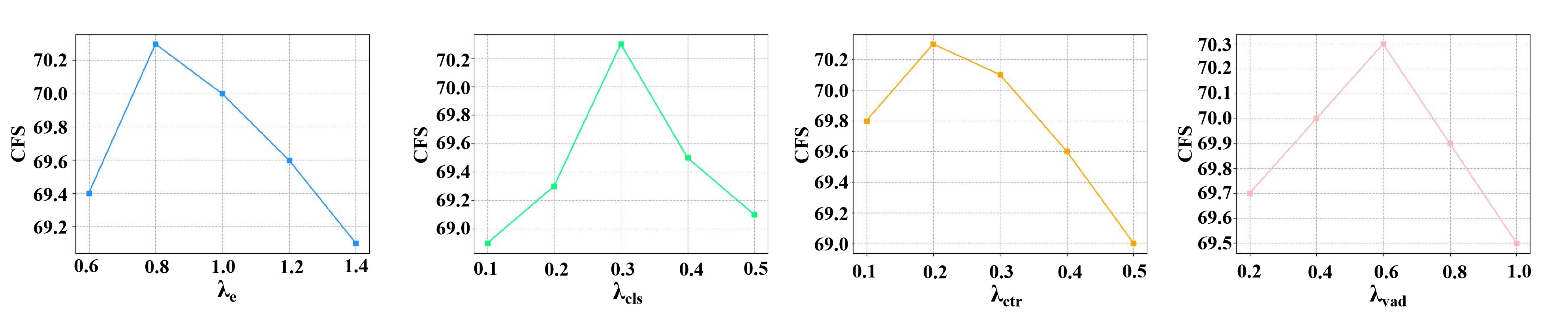}} 
\caption{The effects of three trade-off parameters on EVC-MSVD of $\lambda_e$, $\lambda_{cls}$, $\lambda_{ctr}$, and $\lambda_{vad}$, respectively.}
\label{figloss}
\end{figure*}

\begin{figure*}[t]        
\center{\includegraphics[width=0.9\linewidth] {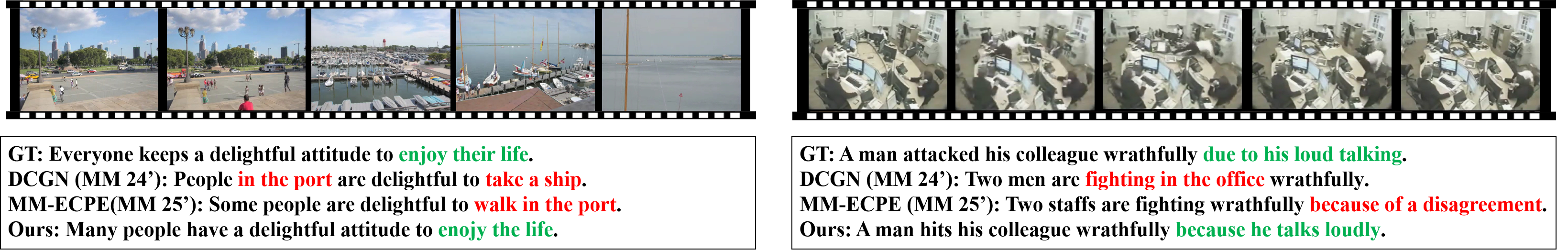}} 
\caption{Qualitative results for comparison between our model and other SOTA methods, \emph{i.e.,} DCGN\cite{ye2024dual} and MM-ECPE\cite{ye2025multi}.}
\label{fig3}
\end{figure*}

\begin{table}[t]
\centering
\caption{\label{human}{The comparison with SOTA methods on human evaluations.}}
\resizebox{0.95\columnwidth}{!}
{
\begin{tabular}{c|cccc}
\toprule
{Metric} & Accuracy&Relevance&Coherence&Usability\\
\midrule
VEIN\cite{song2024emotional}& 5.47 & 5.18 & 6.27 & 5.04 \\
EPAN\cite{song2023emotion}& 6.24 & 5.97 & 7.13 & 5.76\\
DCGN\cite{ye2024dual}& 6.88 & 6.43 & 7.68 & 6.32\\
VideoBLIP\cite{yu2024eliciting}& 5.91& 5.54 & 7.20 & 4.56\\
MM-ECPE\cite{ye2025multi}& 6.40 & 6.75 & 8.04 &7.36 \\
\midrule
GPT-4o\cite{hurst2024gpt}  & 6.98  &6.04  & 7.24  &6.62     \\
VideoLLaMA3\cite{zhang2025videollama}  &7.22   & 6.20 &7.72   & 6.84     \\
{\textbf{Ours}}& \textbf{7.02} & \textbf{6.98} & \textbf{8.44} & \textbf{7.76}\\
\bottomrule
\end{tabular}}
\end{table}

\noindent\textbf{Discussion on proposed modules.} In this paper, we propose two important modules to refine visual and emotional features, named CVSD and VEIL. To further present their effectiveness, we conduct another ablation study shown in Table \ref{module}. We observe that each module could improve its performance independently. Specifically, CVSD refines visual features to locate fine-grained cause concepts, which helps to capture more accurate factual semantics, resulting in obvious improvements on semantic metrics, \emph{i.e.,} +3.4\%/+4.3\% improvements on BLEU-4/CIDEr. Besides, VEIL enhances emotion dictionary features to help capture visual emotion cues and improve the rationality of the emotion-attributed captions, resulting in obvious improvements in emotion accuracy, \emph{i.e.,} +6.1\%/+6.1\% improvements on ${\rm Acc}_{sw}$/${\rm Acc}_{c}$. Finally, we combine two modules to achieve the best performance in both emotional and factual semantics, \emph{i.e.,} +5.8\%/+5.9\% improvements on BFS/CFS.

\noindent\textbf{Discussion on dense components.} In our proposed Concept-aware Visual Semantic Decomposition module, we capture three types of cause concepts to refine the visual features, namely scene, object, and motion concepts. Therefore, it is necessary to discuss the effectiveness of each dense concept for our model. As shown in Table \ref{dense}, we observe that each concept could improve the model's performance independently. Specifically, the scene concept significantly improves emotion accuracy. We analyze that the scene in the video determines the emotional tone. For example, a dark scene often points to negative emotions. Besides, the object and motion concepts significantly improve semantic metrics. They usually include the characters and their actions in the main events of the video, which helps the model accurately generate event-related words. Through combining these three concepts, our model achieves the best performance, demonstrating the superiority of our proposed module.

\noindent\textbf{Discussion on objective functions.} In our framework, in addition to the common emotion classification loss and cross-entropy loss, we also propose two objective functions to optimize our model, namely $\mathcal{L}_{ctr}$ and $\mathcal{L}_{vad}$. We conduct an ablation study to discuss the effectiveness of them. As shown in Table \ref{nce}, we first observe that $\mathcal{L}_{vad}$ could optimize the performance of our model, especially on emotion accuracy, \emph{i.e.,} +2.0\%/+1.8\% improvements on ${\rm Acc}_{sw}$/${\rm Acc}_{c}$. $\mathcal{L}_{vad}$ enables accurate emotional mining by aligning the predicted emotion representation with the vectors in the VAD dictionary. Besides, we discuss the effectiveness of $\mathcal{L}_{ctr}$ under three settings, namely 1st round, 2nd round, and the combination. We observe that applying $\mathcal{L}_{ctr}$ in the 1st round produces only a slight improvement and even hurts some performances on emotion accuracy. Applying $\mathcal{L}_{ctr}$ to ($\widetilde{\mathcal{V}},\ \widetilde{\mathcal{E}}$) may affect the extraction of ($\mathbb{C},\ \mathbb{E}$), causing the emotion and cause features to become homogenized, which hurts the prediction of visual emotions. Besides, compared with the combination setting, applying $\mathcal{L}_{ctr}$ only on the 2nd round achieves better performance.

\noindent\textbf{Discussion on trade-off parameters.}
We introduce several hyper-parameters to control the optimization of our model, namely $\lambda_e$, $\lambda_{cls}$, $\lambda_{ctr}$, and $\lambda_{vad}$. It is essential to discuss the best setting for them. Since the hybrid metric CFS could reflect both emotional accuracy and factual semantics, we evaluate the performance of different parameter settings on it. As shown in Figure \ref{figloss}, the model achieves the best performance when $\lambda_e=0.8$, $\lambda_{cls}=0.3$, $\lambda_{ctr}=0.2$, and $\lambda_{vad}=0.6$, which is consistent with our setting.

\subsection{Human Evaluation}
Due to the highly subjective nature of the EVC task, it is a crucial criterion for evaluation whether the generated descriptions conform to human preferences. Therefore, we add human evaluation to fully measure the quality of emotional descriptions. Inspired by the work \cite{wang2021emotion}, we designs four metrics, including (1) emotion accuracy: assess the accuracy of emotional expression, (2) relevance: assess whether the generated description is relevant to the video content, (3) coherence: assess the coherence and readability of the description, and (4) usability: how useful would the description be for a person (especially a blind person) to understand what is happening in the video. The score of each metric ranges from 1 to 10. Subsequently, we invite 20 participants with excellent English skills to score a subset of 50 video-caption pairs on these four metrics and calculate the average score on each metric. We make a comparison with two types of models on human evaluation: the state-of-the-art models on EVC and several large vision-language models (LVLMs).

\noindent\textbf{Comparison with SOTA methods. }As shown in Table \ref{human}, compared with the SOTA methods for EVC, our model is significantly ahead in all four metrics. Our proposed CVSD captures fine-grained cause concepts, thereby generating more accurate and diverse vocabulary. Furthermore, our proposed VEIL utilizes VAD constraints to enhance the interpretability of emotions, making the generated descriptions more easily empathized with and understood by humans. These show that our model does not simply imitate the reference sentences, but learns the representation of emotional and factual semantics.

\noindent\textbf{Comparison with LVLMs. }As shown in Table \ref{human}, despite the outstanding performance of LVLMs, our model still outperforms them in human evaluations. Due to their powerful ability in objective understanding and generation, LVLMs could produce a wide variety of descriptions. However, limited by insufficient emotion understanding and the influence of visual hallucinations, they sometimes generate factual and emotional contents irrelevant to the video, leading to lower human ratings. Conversely, our model filters out emotion-irrelevant visual information and visual-irrelevant emotional information through fine-grained visual and emotional refinement, thereby generating emotional descriptions that are more closely aligned with the video content.

\begin{figure*}[t]        
\center{\includegraphics[width=0.8\linewidth] {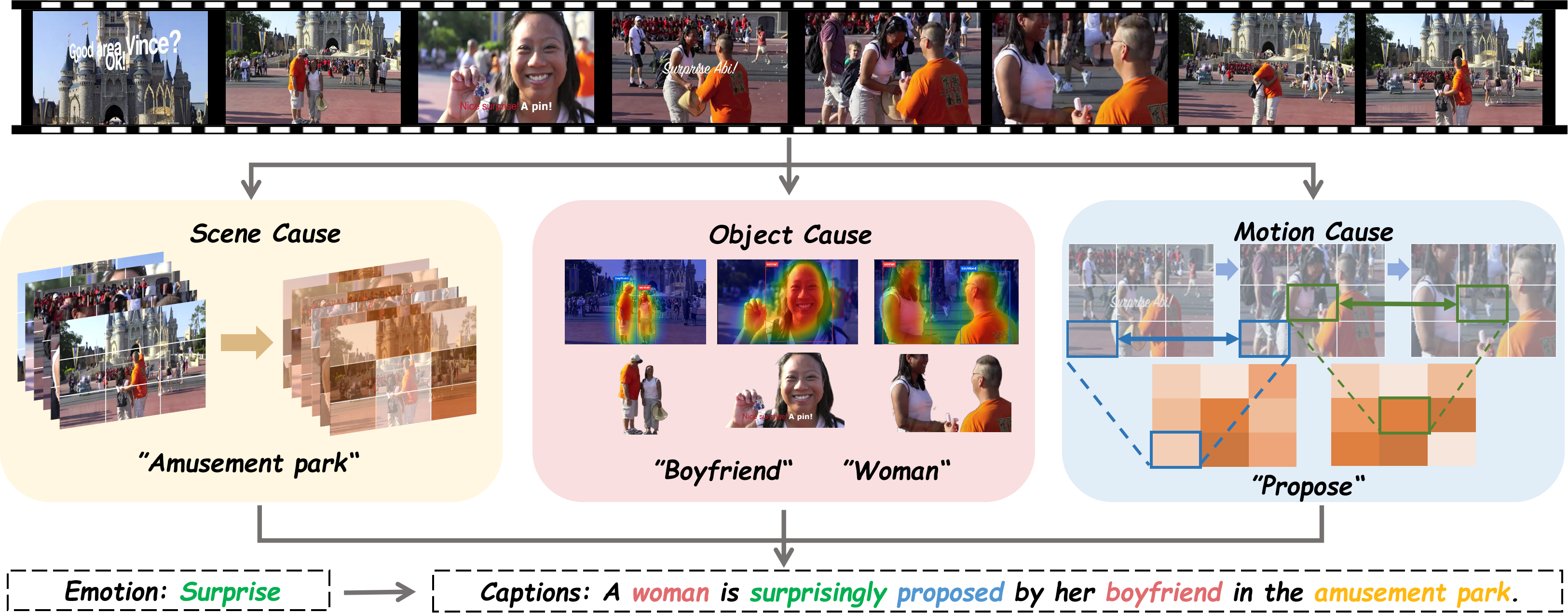}} 
\caption{Qualitative results for our proposed Concept-aware Visual Semantic Decomposition module. The yellow, red, and blue words in the caption correspond to the scene, object, and motion cause exploration.}
\label{fig4}
\end{figure*}

\begin{figure}[t]        
\center{\includegraphics[width=0.9\linewidth] {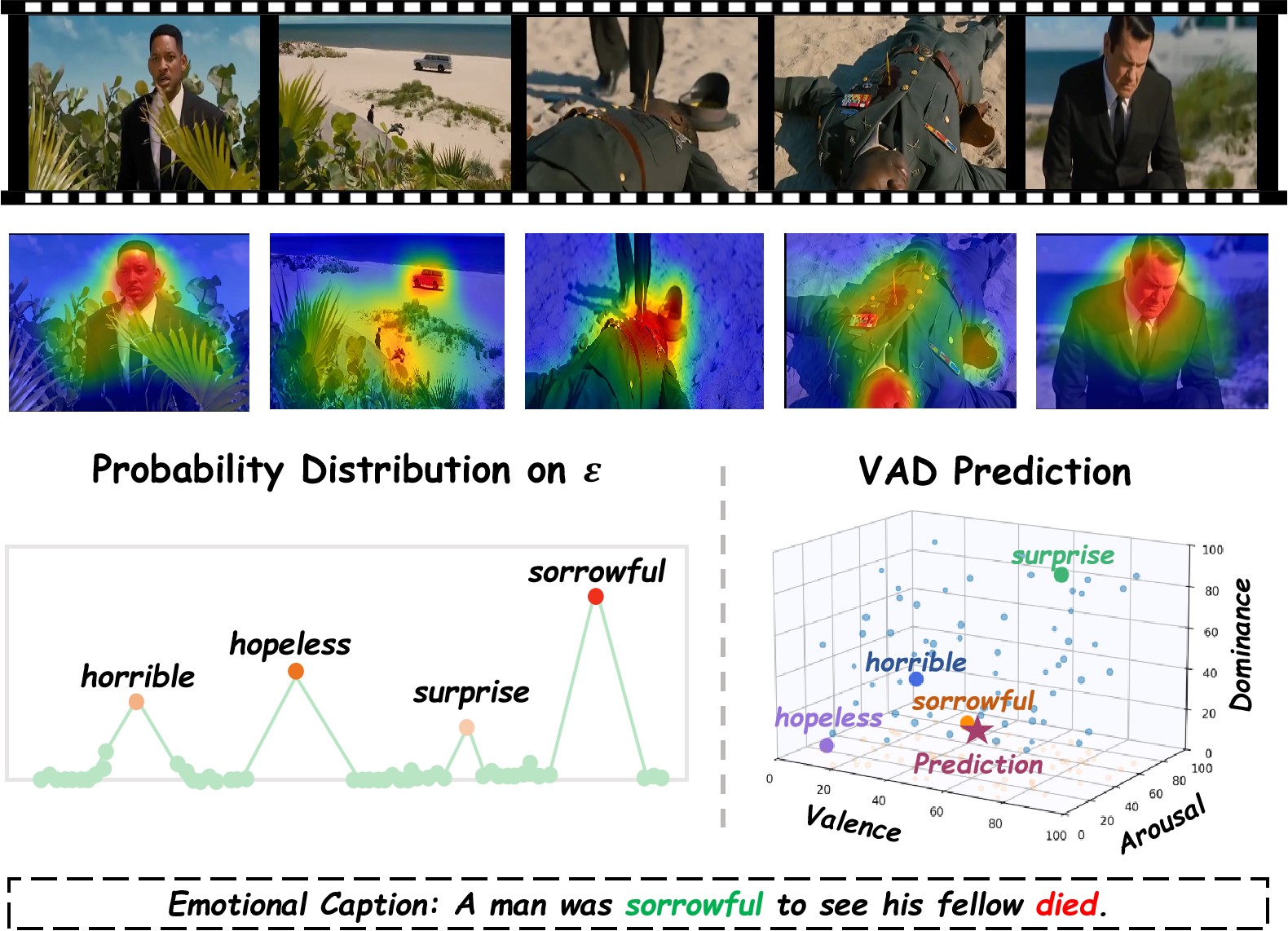}} 
\caption{Qualitative results for our proposed Visual-guided Emotion Interpretable Learning module.}
\label{fig5}
\end{figure}

\subsection{Qualitative Results}
To intuitively demonstrate the advantage of our model, we make a visualization from several aspects: comparison with the state-of-the-art methods, visualization of proposed CVSD, and visualization of proposed VEIL.

\noindent\textbf{Comparison with the SOTA methods.} As shown in Fig. \ref{fig3}, compared with the SOTA methods DCGN\cite{ye2024dual} and MM-ECPE\cite{ye2025multi}, our method generates more accurate emotional captions in both two cases. Firstly, we observe that cause extraction is helpful to locate more accurate emotion causes, \emph{i.e.,} ``enjoy the life'' and ``his loud talking''. Compared with global visual features, fine-granularity visual features could focus on emotion-related visual regions and capture more detailed factual semantics. Besides, we observe that emotion-cause pair collaborative extraction could also improve emotion perception, which results in correct emotion word prediction ``delightful'' and ``wrathfully''. Specifically, our model accurately captures emotion-related cause concepts and applies it for emotion perception, which filters out the emotion errors caused by irrelevant visual noises. 

\noindent\textbf{Visualization of proposed CVSD.} To demonstrate the effectiveness of our proposed CVSD module, we make a visualization for the extraction of different concepts (\emph{e.g.,} scene, object, and motion). As shown in Fig. \ref{fig4}, the proposed CVSD accurately captures the different types of semantic concepts corresponding to specific words in the generated sentence. For the scene cause, the global attention forces our model to focus on the background and environment in the video, resulting in the generation of the word \textit{``amusement park''}. For the object cause, we employ spatial-geometry and temporal-block attention to focus on the main character, generating the key words \textit{``boyfriend''} and \textit{``woman''}. For the motion cause, our model focuses on the dynamic changes of key visual regions over temporal dimension, resulting in the correct word \textit{``propose''}. Overall, the proposed CVSD achieves accurate mining of potential emotional causes by decoupling different types of semantic concepts in videos, providing a credible foundation for mining visual emotions.

\noindent\textbf{Visualization of proposed VEIL.} To intuitively demonstrate the process of capturing visual emotions, we make a visualization for our proposed VEIL module. As shown in Fig. \ref{fig5}, the proposed VEIL improves the accuracy of emotion mining by two constraints, including probability distribution in the emotion dictionary and interpretable VAD vector prediction. Firstly, we observe that in the distribution on $\mathcal{E}$, the correct emotion word \textit{``sorrowful''} has the highest probability. Besides, in the VAD semantic space, our prediction VAD vector achieves the closest space distance to the vector of \textit{``sorrowful''}. The results above demonstrate that our proposed VEIL could not only obtain the optimal emotion distribution through cross-attention between video and emotion dictionary, but also analyze video content from the aspects of arousal and valence under the constraint of VAD knowledge, thereby improving the interpretability of emotion mining.
\section{Conclusion} 
In this paper, we propose MM-ECPE++, a novel Fine-grained Emotion-Cause Pair Extraction framework for Emotion-Attributed Video Captioning, which performs pairwise extraction of emotional cues and visual causes through multi-round step-by-step refinement. The pairwise extraction promotes each other mutually, enhancing the accuracy and interpretability of emotion perception and helping to decouple the factual and emotional description generation. To be specific, in the 1st round, we propose a concept-aware visual semantic decomposition to refine the visual features through fine-grained concepts mining, including scene, object, and motion concepts. Besides, we propose a visual-guided emotion interpretable learning to filter visual-irrelevant emotional information through visual guiding and VAD-vector constraint. Subsequently, we achieve emotion-cause pair extraction in the 2nd round by cross-coupling the visual and emotional features before and after refinement, and leveraging contrastive loss to achieve semantic forced alignment. Finally, we aggregate visual and emotion-cause semantics into the LLMs-based decoder through a Q-former component for more accurate and vivid emotional captions. Extensive experiments on challenging datasets demonstrate the superiority of our method and each proposed module.

\bibliographystyle{IEEEtran}
\bibliography{sample-base}

\begin{thebibliography}{10}
\providecommand{\url}[1]{#1}
\csname url@samestyle\endcsname
\providecommand{\newblock}{\relax}
\providecommand{\bibinfo}[2]{#2}
\providecommand{\BIBentrySTDinterwordspacing}{\spaceskip=0pt\relax}
\providecommand{\BIBentryALTinterwordstretchfactor}{4}
\providecommand{\BIBentryALTinterwordspacing}{\spaceskip=\fontdimen2\font plus
\BIBentryALTinterwordstretchfactor\fontdimen3\font minus \fontdimen4\font\relax}
\providecommand{\BIBforeignlanguage}[2]{{%
\expandafter\ifx\csname l@#1\endcsname\relax
\typeout{** WARNING: IEEEtran.bst: No hyphenation pattern has been}%
\typeout{** loaded for the language `#1'. Using the pattern for}%
\typeout{** the default language instead.}%
\else
\language=\csname l@#1\endcsname
\fi
#2}}
\providecommand{\BIBdecl}{\relax}
\BIBdecl

\bibitem{wang2025mdkat}
J.~Wang, C.~Wang, L.~Guo, S.~Zhao, D.~Wang, S.~Zhang, X.~Zhao, J.~Yu, Y.~Wang, Y.~Yang \emph{et~al.}, ``Mdkat: Multimodal decoupling with knowledge aggregation and transfer for video emotion recognition,'' \emph{IEEE Transactions on Circuits and Systems for Video Technology}, 2025.

\bibitem{li2025feature}
S.~Li, C.~Lu, Y.~Zong, H.~Lian, and W.~Zheng, ``Feature evaluation and joint interaction for audio-visual emotion recognition,'' \emph{IEEE Transactions on Circuits and Systems for Video Technology}, 2025.

\bibitem{pennington2014glove}
J.~Pennington, R.~Socher, and C.~D. Manning, ``Glove: Global vectors for word representation,'' in \emph{Proceedings of the 2014 conference on empirical methods in natural language processing (EMNLP)}, 2014, pp. 1532--1543.

\bibitem{chen2023weakly}
W.~Chen, G.~Li, X.~Zhang, S.~Wang, L.~Li, and Q.~Huang, ``Weakly supervised text-based actor-action video segmentation by clip-level multi-instance learning,'' \emph{ACM Transactions on Multimedia Computing, Communications and Applications}, vol.~19, no.~1, pp. 1--22, 2023.

\bibitem{huang2025graph}
X.~Huang, W.~Chen, B.~Hu, and Z.~Mao, ``Graph mixture of experts and memory-augmented routers for multivariate time series anomaly detection,'' in \emph{Proceedings of the AAAI Conference on Artificial Intelligence}, vol.~39, no.~16, 2025, pp. 17\,476--17\,484.

\bibitem{hurst2024gpt}
A.~Hurst, A.~Lerer, A.~P. Goucher, A.~Perelman, A.~Ramesh, A.~Clark, A.~Ostrow, A.~Welihinda, A.~Hayes, A.~Radford \emph{et~al.}, ``Gpt-4o system card,'' \emph{arXiv preprint arXiv:2410.21276}, 2024.

\bibitem{li2022ecpec}
W.~Li, Y.~Li, V.~Pandelea, M.~Ge, L.~Zhu, and E.~Cambria, ``Ecpec: Emotion-cause pair extraction in conversations,'' \emph{IEEE Transactions on Affective Computing}, vol.~14, no.~3, pp. 1754--1765, 2022.

\bibitem{ye2025multi}
C.~Ye, W.~Chen, P.~Song, X.~Liu, L.~Zhang, and Z.~Mao, ``Multi-round mutual emotion-cause pair extraction for emotion-attributed video captioning,'' in \emph{Proceedings of the 33rd ACM International Conference on Multimedia}, 2025, pp. 3320--3329.

\bibitem{an2023global}
J.~An, Z.~Ding, K.~Li, and R.~Xia, ``Global-view and speaker-aware emotion cause extraction in conversations,'' \emph{IEEE/ACM Transactions on Audio, Speech, and Language Processing}, vol.~31, pp. 3814--3823, 2023.

\bibitem{li2024multimodal}
B.~Li, H.~Fei, F.~Li, T.-s. Chua, and D.~Ji, ``Multimodal emotion-cause pair extraction with holistic interaction and label constraint,'' \emph{ACM Transactions on Multimedia Computing, Communications and Applications}, 2024.

\bibitem{wang2018reconstruction}
B.~Wang, L.~Ma, W.~Zhang, and W.~Liu, ``Reconstruction network for video captioning,'' in \emph{Proceedings of the IEEE conference on computer vision and pattern recognition}, 2018, pp. 7622--7631.

\bibitem{kingma2014adam}
D.~P. Kingma and J.~Ba, ``Adam: A method for stochastic optimization,'' \emph{arXiv preprint arXiv:1412.6980}, 2014.

\bibitem{wang2024enhancing}
B.~Wang, K.~Tang, and P.~Zhu, ``Enhancing emotion-cause pair extraction in conversations via center event detection and reasoning,'' in \emph{Findings of the Association for Computational Linguistics: EMNLP 2024}, 2024, pp. 10\,773--10\,783.

\bibitem{yu2024prompting}
T.~Yu, K.~Fu, S.~Wang, Q.~Huang, and J.~Yu, ``Prompting video-language foundation models with domain-specific fine-grained heuristics for video question answering,'' \emph{IEEE Transactions on Circuits and Systems for Video Technology}, vol.~35, no.~2, pp. 1615--1630, 2024.

\bibitem{banerjee2005meteor}
S.~Banerjee and A.~Lavie, ``Meteor: An automatic metric for mt evaluation with improved correlation with human judgments,'' in \emph{Proceedings of the acl workshop on intrinsic and extrinsic evaluation measures for machine translation and/or summarization}, 2005, pp. 65--72.

\bibitem{hu2021loralowrankadaptationlarge}
\BIBentryALTinterwordspacing
E.~J. Hu, Y.~Shen, P.~Wallis, Z.~Allen-Zhu, Y.~Li, S.~Wang, L.~Wang, and W.~Chen, ``Lora: Low-rank adaptation of large language models,'' 2021. [Online]. Available: \url{https://arxiv.org/abs/2106.09685}
\BIBentrySTDinterwordspacing

\bibitem{raffel2020exploring}
C.~Raffel, N.~Shazeer, A.~Roberts, K.~Lee, S.~Narang, M.~Matena, Y.~Zhou, W.~Li, and P.~J. Liu, ``Exploring the limits of transfer learning with a unified text-to-text transformer,'' \emph{Journal of machine learning research}, vol.~21, no. 140, pp. 1--67, 2020.

\bibitem{gao2025learning}
J.~Gao, M.~Chen, and C.~Xu, ``Learning probabilistic presence-absence evidence for weakly-supervised audio-visual event perception,'' \emph{IEEE Transactions on Pattern Analysis and Machine Intelligence}, 2025.

\bibitem{lan2025expllm}
X.~Lan, J.~Xue, J.~Qi, D.~Jiang, K.~Lu, and T.-S. Chua, ``Expllm: Towards chain of thought for facial expression recognition,'' \emph{IEEE Transactions on Multimedia}, 2025.

\bibitem{guo2024benchmarking}
D.~Guo, K.~Li, B.~Hu, Y.~Zhang, and M.~Wang, ``Benchmarking micro-action recognition: Dataset, method, and application,'' \emph{IEEE Transactions on Circuits and Systems for Video Technology}, 2024.

\bibitem{song2022contextual}
P.~Song, D.~Guo, J.~Cheng, and M.~Wang, ``Contextual attention network for emotional video captioning,'' \emph{IEEE Transactions on Multimedia}, 2022.

\bibitem{wang2024observe}
F.~Wang, H.~Ma, X.~Shen, J.~Yu, and R.~Xia, ``Observe before generate: Emotion-cause aware video caption for multimodal emotion cause generation in conversations,'' in \emph{Proceedings of the 32nd ACM International Conference on Multimedia}, 2024, pp. 5820--5828.

\bibitem{radford2021learning}
A.~Radford, J.~W. Kim, C.~Hallacy, A.~Ramesh, G.~Goh, S.~Agarwal, G.~Sastry, A.~Askell, P.~Mishkin, J.~Clark \emph{et~al.}, ``Learning transferable visual models from natural language supervision,'' in \emph{International conference on machine learning}.\hskip 1em plus 0.5em minus 0.4em\relax PMLR, 2021, pp. 8748--8763.

\bibitem{xu2024cross}
N.~Xu, Y.~Gao, T.-T. Zhang, H.~Tian, and A.-A. Liu, ``Cross-modal coherence-enhanced feedback prompting for news captioning,'' in \emph{Proceedings of the 32nd ACM International Conference on Multimedia}, 2024, pp. 9369--9377.

\bibitem{vedantam2015cider}
R.~Vedantam, C.~Lawrence~Zitnick, and D.~Parikh, ``Cider: Consensus-based image description evaluation,'' in \emph{Proceedings of the IEEE conference on computer vision and pattern recognition}, 2015, pp. 4566--4575.

\bibitem{ryu2021semantic}
H.~Ryu, S.~Kang, H.~Kang, and C.~D. Yoo, ``Semantic grouping network for video captioning,'' in \emph{proceedings of the AAAI Conference on Artificial Intelligence}, vol.~35, no.~3, 2021, pp. 2514--2522.

\bibitem{xu2024rule}
N.~Xu, T.~Zhang, H.~Tian, and A.-A. Liu, ``Rule-driven news captioning,'' \emph{IEEE Transactions on Circuits and Systems for Video Technology}, 2024.

\bibitem{yu2024eliciting}
K.~Yu, Z.~Zhang, F.~Hu, S.~Storks, and J.~Chai, ``Eliciting in-context learning in vision-language models for videos through curated data distributional properties,'' in \emph{Proceedings of the 2024 Conference on Empirical Methods in Natural Language Processing}, 2024, pp. 20\,416--20\,431.

\bibitem{qi2024versatile}
F.~Qi, H.~Zhang, X.~Yang, and C.~Xu, ``A versatile multimodal learning framework for zero-shot emotion recognition,'' \emph{IEEE Transactions on Circuits and Systems for Video Technology}, vol.~34, no.~7, pp. 5728--5741, 2024.

\bibitem{chen2021cascade}
W.~Chen, G.~Li, X.~Zhang, H.~Yu, S.~Wang, and Q.~Huang, ``Cascade cross-modal attention network for video actor and action segmentation from a sentence,'' in \emph{Proceedings of the 29th ACM International Conference on Multimedia}, 2021, pp. 4053--4062.

\bibitem{xia2019emotion}
R.~Xia and Z.~Ding, ``Emotion-cause pair extraction: A new task to emotion analysis in texts,'' in \emph{Proceedings of the 57th Annual Meeting of the Association for Computational Linguistics}, 2019, pp. 1003--1012.

\bibitem{chen2011collecting}
D.~Chen and W.~B. Dolan, ``Collecting highly parallel data for paraphrase evaluation,'' in \emph{Proceedings of the 49th annual meeting of the association for computational linguistics: human language technologies}, 2011, pp. 190--200.

\bibitem{chen2024coarse}
X.~Chen, C.~Yang, C.~Sun, M.~Lan, and A.~Zhou, ``From coarse to fine: A distillation method for fine-grained emotion-causal span pair extraction in conversation,'' in \emph{Proceedings of the AAAI Conference on Artificial Intelligence}, vol.~38, no.~16, 2024, pp. 17\,790--17\,798.

\bibitem{ma2024extraction}
H.~Ma, J.~Yu, F.~Wang, H.~Cao, and R.~Xia, ``From extraction to generation: multimodal emotion-cause pair generation in conversations,'' \emph{IEEE Transactions on Affective Computing}, 2024.

\bibitem{wang2023improving}
T.~Wang, W.~Chen, Y.~Tian, Y.~Song, and Z.~Mao, ``Improving image captioning via predicting structured concepts,'' in \emph{Proceedings of the 2023 Conference on Empirical Methods in Natural Language Processing}, 2023, pp. 360--370.

\bibitem{liu2024bootstrapping}
C.~Liu, Y.~Tian, W.~Chen, Y.~Song, and Y.~Zhang, ``Bootstrapping large language models for radiology report generation,'' in \emph{Proceedings of the AAAI Conference on Artificial Intelligence}, vol.~38, no.~17, 2024, pp. 18\,635--18\,643.

\bibitem{zhang2025videollama}
B.~Zhang, K.~Li, Z.~Cheng, Z.~Hu, Y.~Yuan, G.~Chen, S.~Leng, Y.~Jiang, H.~Zhang, X.~Li \emph{et~al.}, ``Videollama 3: Frontier multimodal foundation models for image and video understanding,'' \emph{arXiv preprint arXiv:2501.13106}, 2025.

\bibitem{jin2024improving}
Y.~Jin, W.~Chen, Y.~Tian, Y.~Song, C.~Yan, and Z.~Mao, ``Improving radiology report generation with d 2-net: When diffusion meets discriminator,'' in \emph{ICASSP 2024-2024 IEEE International Conference on Acoustics, Speech and Signal Processing (ICASSP)}.\hskip 1em plus 0.5em minus 0.4em\relax IEEE, 2024, pp. 2215--2219.

\bibitem{jin2024improving2}
Y.~Jin, W.~Chen, Y.~Tian, Y.~Song, and C.~Yan, ``Improving radiology report generation with multi-grained abnormality prediction,'' \emph{Neurocomputing}, vol. 600, p. 128122, 2024.

\bibitem{liu2025enriched}
A.-A. Liu, Q.~Wu, N.~Xu, H.~Tian, and L.~Wang, ``Enriched image captioning based on knowledge divergence and focus,'' \emph{IEEE Transactions on Circuits and Systems for Video Technology}, 2025.

\bibitem{song2024emotional}
P.~Song, D.~Guo, X.~Yang, S.~Tang, and M.~Wang, ``Emotional video captioning with vision-based emotion interpretation network,'' \emph{IEEE Transactions on Image Processing}, 2024.

\bibitem{song2023emotion}
P.~Song, D.~Guo, X.~Yang, S.~Tang, E.~Yang, and M.~Wang, ``Emotion-prior awareness network for emotional video captioning,'' in \emph{Proceedings of the 31st ACM International Conference on Multimedia}, 2023, pp. 589--600.

\bibitem{wang2025combatting}
C.~Wang, W.~Chen, X.~Cui, Y.~Zhao, Z.~Qi, P.~Huang, X.~Liu, and W.~Zhang, ``Combatting data imbalance and noise in micro-action recognition,'' in \emph{Proceedings of the 33rd ACM International Conference on Multimedia}, 2025, pp. 14\,229--14\,235.

\bibitem{yu-etal-2024-eliciting}
\BIBentryALTinterwordspacing
K.~Yu, Z.~Zhang, F.~Hu, S.~Storks, and J.~Chai, ``Eliciting in-context learning in vision-language models for videos through curated data distributional properties,'' in \emph{Proceedings of the 2024 Conference on Empirical Methods in Natural Language Processing}, Y.~Al-Onaizan, M.~Bansal, and Y.-N. Chen, Eds.\hskip 1em plus 0.5em minus 0.4em\relax Miami, Florida, USA: Association for Computational Linguistics, Nov. 2024, pp. 20\,416--20\,431. [Online]. Available: \url{https://aclanthology.org/2024.emnlp-main.1137}
\BIBentrySTDinterwordspacing

\bibitem{mohammad2018obtaining}
S.~Mohammad, ``Obtaining reliable human ratings of valence, arousal, and dominance for 20,000 english words,'' in \emph{Proceedings of the 56th annual meeting of the association for computational linguistics (volume 1: Long papers)}, 2018, pp. 174--184.

\bibitem{fu2024linguistic}
Z.~Fu, L.~Zhang, H.~Xia, and Z.~Mao, ``Linguistic-aware patch slimming framework for fine-grained cross-modal alignment,'' in \emph{Proceedings of the IEEE/CVF Conference on Computer Vision and Pattern Recognition}, 2024, pp. 26\,307--26\,316.

\bibitem{wang2025emotion}
Y.~Wang, Y.~Liu, S.~Zhou, Y.~Huang, C.~Tang, W.~Zhou, and Z.~Chen, ``Emotion-oriented cross-modal prompting and alignment for human-centric emotional video captioning,'' \emph{IEEE Transactions on Multimedia}, 2025.

\bibitem{ye2024dual}
C.~Ye, W.~Chen, J.~Li, L.~Zhang, and Z.~Mao, ``Dual-path collaborative generation network for emotional video captioning,'' in \emph{Proceedings of the 32nd ACM International Conference on Multimedia}, 2024, p. 496–505.

\bibitem{lin2004rouge}
C.-Y. Lin, ``Rouge: A package for automatic evaluation of summaries,'' in \emph{Text summarization branches out}, 2004, pp. 74--81.

\bibitem{zhu2024knowledge}
P.~Zhu, B.~Wang, K.~Tang, H.~Zhang, X.~Cui, and Z.~Wang, ``A knowledge-guided graph attention network for emotion-cause pair extraction,'' \emph{Knowledge-Based Systems}, vol. 286, p. 111342, 2024.

\bibitem{yu2023comprehensive}
T.~Yu, X.~Lin, S.~Wang, W.~Sheng, Q.~Huang, and J.~Yu, ``A comprehensive survey of 3d dense captioning: Localizing and describing objects in 3d scenes,'' \emph{IEEE Transactions on Circuits and Systems for Video Technology}, vol.~34, no.~3, pp. 1322--1338, 2023.

\bibitem{jiang2014predicting}
Y.-G. Jiang, B.~Xu, and X.~Xue, ``Predicting emotions in user-generated videos,'' in \emph{Proceedings of the AAAI conference on artificial intelligence}, vol.~28, no.~1, 2014.

\bibitem{chen2022multi}
W.~Chen, D.~Hong, Y.~Qi, Z.~Han, S.~Wang, L.~Qing, Q.~Huang, and G.~Li, ``Multi-attention network for compressed video referring object segmentation,'' in \emph{Proceedings of the 30th ACM International Conference on Multimedia}, 2022, pp. 4416--4425.

\bibitem{song2025towards}
P.~Song, L.~Zhang, L.~Lan, W.~Chen, D.~Guo, X.~Yang, and M.~Wang, ``Towards efficient partially relevant video retrieval with active moment discovering,'' \emph{IEEE Transactions on Multimedia}, 2025.

\bibitem{gao2023vectorized}
J.~Gao, M.~Chen, and C.~Xu, ``Vectorized evidential learning for weakly-supervised temporal action localization,'' \emph{IEEE Transactions on Pattern Analysis and Machine Intelligence}, vol.~45, no.~12, pp. 15\,949 -- 15\,963, 2023.

\bibitem{fu2024sentiment}
F.~Fu, S.~Fang, W.~Chen, and Z.~Mao, ``Sentiment-oriented transformer-based variational autoencoder network for live video commenting,'' \emph{ACM Transactions on Multimedia Computing, Communications and Applications}, vol.~20, no.~4, pp. 1--24, 2024.

\bibitem{lin2024prompting}
Z.~Lin, W.~Chen, Y.~Song, and Y.~Zhang, ``Prompting few-shot multi-hop question generation via comprehending type-aware semantics,'' in \emph{Findings of the Association for Computational Linguistics: NAACL 2024}, 2024, pp. 3730--3740.

\bibitem{fard2025affectnet+}
A.~P. Fard, M.~M. Hosseini, T.~D. Sweeny, and M.~H. Mahoor, ``Affectnet+: A database for enhancing facial expression recognition with soft-labels,'' \emph{IEEE Transactions on Affective Computing}, 2025.

\bibitem{wang2021emotion}
H.~Wang, P.~Tang, Q.~Li, and M.~Cheng, ``Emotion expression with fact transfer for video description,'' \emph{IEEE Transactions on Multimedia}.

\bibitem{tang2021graph}
S.~Tang, D.~Guo, R.~Hong, and M.~Wang, ``Graph-based multimodal sequential embedding for sign language translation,'' \emph{IEEE Transactions on Multimedia}, vol.~24, pp. 4433--4445, 2021.

\bibitem{10756281}
H.~Li, X.~Liu, G.~Li, S.~Wang, L.~Qing, and Q.~Huang, ``Boost tracking by natural language with prompt-guided grounding,'' \emph{IEEE Transactions on Intelligent Transportation Systems}, vol.~26, no.~1, pp. 1088--1100, 2025.

\bibitem{wang2022multimodal}
F.~Wang, Z.~Ding, R.~Xia, Z.~Li, and J.~Yu, ``Multimodal emotion-cause pair extraction in conversations,'' \emph{IEEE Transactions on Affective Computing}, vol.~14, no.~3, pp. 1832--1844, 2022.

\bibitem{papineni2002bleu}
K.~Papineni, S.~Roukos, T.~Ward, and W.-J. Zhu, ``Bleu: a method for automatic evaluation of machine translation,'' in \emph{Proceedings of the 40th annual meeting of the Association for Computational Linguistics}, 2002, pp. 311--318.

\bibitem{deng2021syntax}
J.~Deng, L.~Li, B.~Zhang, S.~Wang, Z.~Zha, and Q.~Huang, ``Syntax-guided hierarchical attention network for video captioning,'' \emph{IEEE Transactions on Circuits and Systems for Video Technology}, vol.~32, no.~2, pp. 880--892, 2021.

\bibitem{ju2025enhanced}
X.~Ju, D.~Zhang, J.~Li, S.~Li, and G.~Zhou, ``Enhanced generative framework with llms for multimodal emotion-cause pair extraction in conversations,'' \emph{IEEE Transactions on Multimedia}, 2025.

\bibitem{ye2025improving}
C.~Ye, W.~Chen, B.~Hu, L.~Zhang, Y.~Zhang, and Z.~Mao, ``Improving video summarization by exploring the coherence between corresponding captions,'' \emph{IEEE Transactions on Image Processing}, 2025.

\bibitem{hu2023emotion}
G.~Hu, Y.~Zhao, and G.~Lu, ``Emotion prediction oriented method with multiple supervisions for emotion-cause pair extraction,'' \emph{IEEE/ACM Transactions on Audio, Speech, and Language Processing}, vol.~31, pp. 1141--1152, 2023.

\bibitem{chen2026subjective}
W.~Chen, C.~Ye, P.~Song, L.~Zhang, Y.~Zhang, and Z.~Mao, ``Subjective-objective emotion correlated generation network for subjective video captioning,'' \emph{IEEE Transactions on Image Processing}, 2026.

\end{thebibliography}

\begin{IEEEbiography}[{\includegraphics[width=1in,height=1.25in,clip,keepaspectratio]{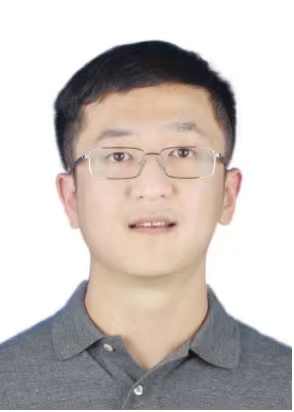}}]{Weidong Chen} (member, IEEE) received the Ph.D. degree in computer application technology from University of Chinese Academy of Sciences, in 2022. He is currently an Associate Researcher with the School of Information Science and Technology, University of Science and Technology of China, Hefei, China. He was a post-doctor with the School of Information Science and Technology, University of Science and Technology of China, from 2022 to 2024. His research interests include computer vision, natural language processing and cross-modal understanding.
\end{IEEEbiography}

\begin{IEEEbiography}[{\includegraphics[width=1in,height=1.25in,clip,keepaspectratio]{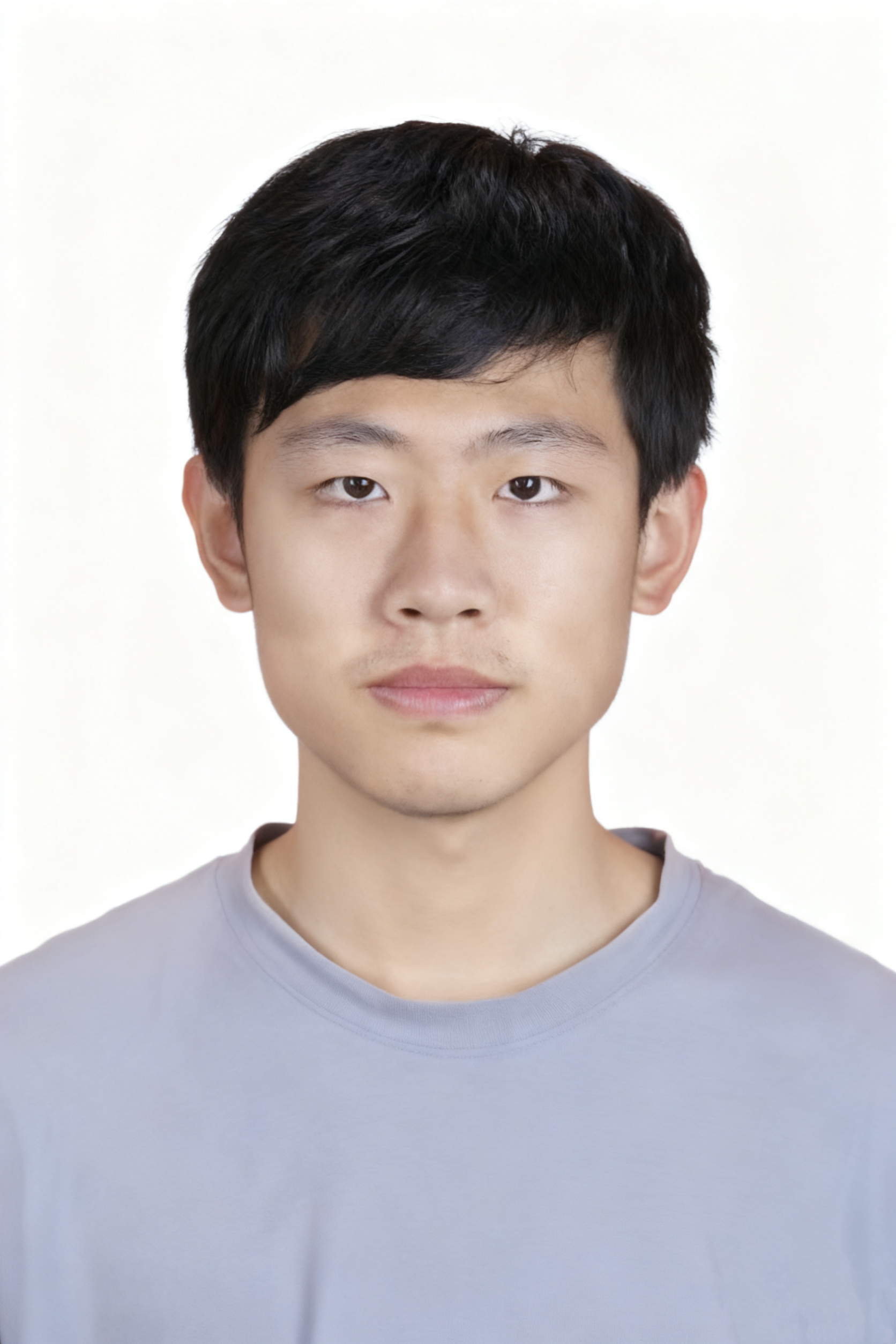}}]{Cheng Ye} received the B.E. degree in the School of Cyberspace Science and Technology, University of Science and Technology of China, Hefei, China, in 2024. He is currently a master student with the School of Information Science and Technology, University of Science and Technology of China, Hefei, China. His research interests include emotional intelligence, video analysis, and multimodal understanding.
\end{IEEEbiography}

\begin{IEEEbiography}[{\includegraphics[width=1in,height=1.25in,clip,keepaspectratio]{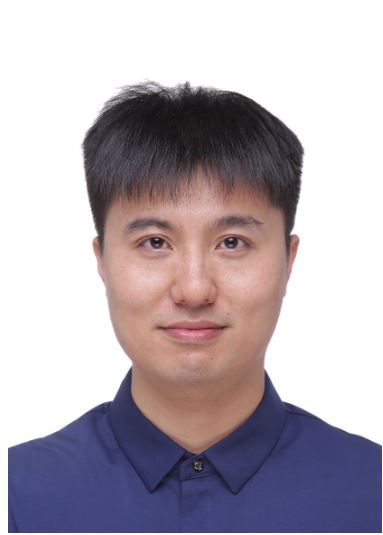}}]{Zhendong Mao} received the Ph.D. degree in computer application technology from the Institute of Computing Technology, Chinese Academy of Sciences, in 2014. He is currently a professor with the School of Cyberspace Science and Technology, University of Science and Technology of China, Hefei, China. He was an assistant professor with the Institute of Information Engineering, Chinese Academy of Sciences, Beijing, from 2014 to 2018. His research interests include cross-modal understanding and cross-modal generation. He serves as an Associate Editor of the IEEE TRANSACTIONS ON CIRCUITS AND SYSTEMS FOR VIDEO TECHNOLOGY and IEEE TRANSACTIONS ON MULTIMEDIA.
\end{IEEEbiography}

\begin{IEEEbiography}[{\includegraphics[width=1in,height=1.25in,clip,keepaspectratio]{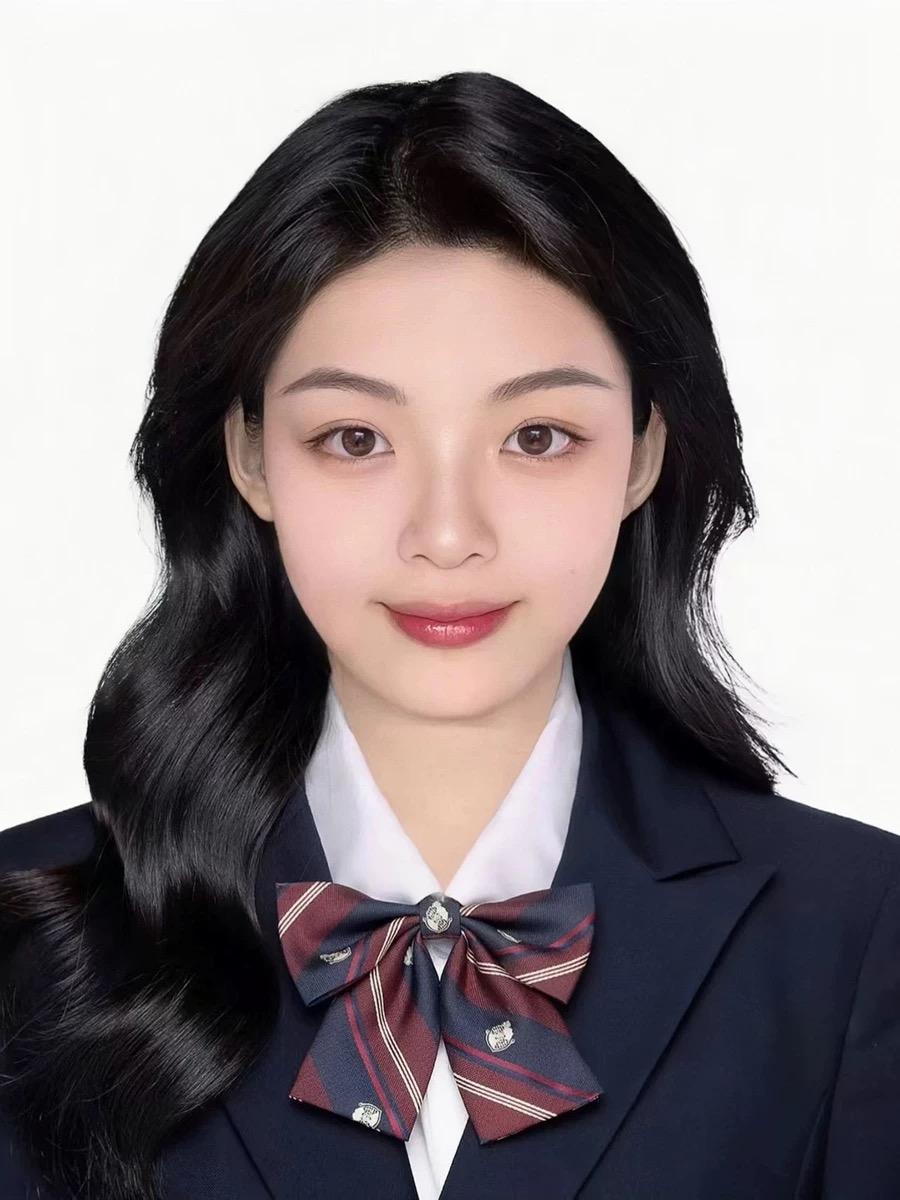}}]{Liping Wang} received the B.E. degree in Information Engineering from Sun Yat-sen University, Guangzhou, China. She is currently a master student with the Institute of Advanced Technology, University of Science and Technology of China, Hefei, China. Her research interests include multimodal empathetic dialogue generation, multi-agent systems, and affective computing.
\end{IEEEbiography}

\begin{IEEEbiography}[{\includegraphics[width=1in,height=1.25in,clip,keepaspectratio]{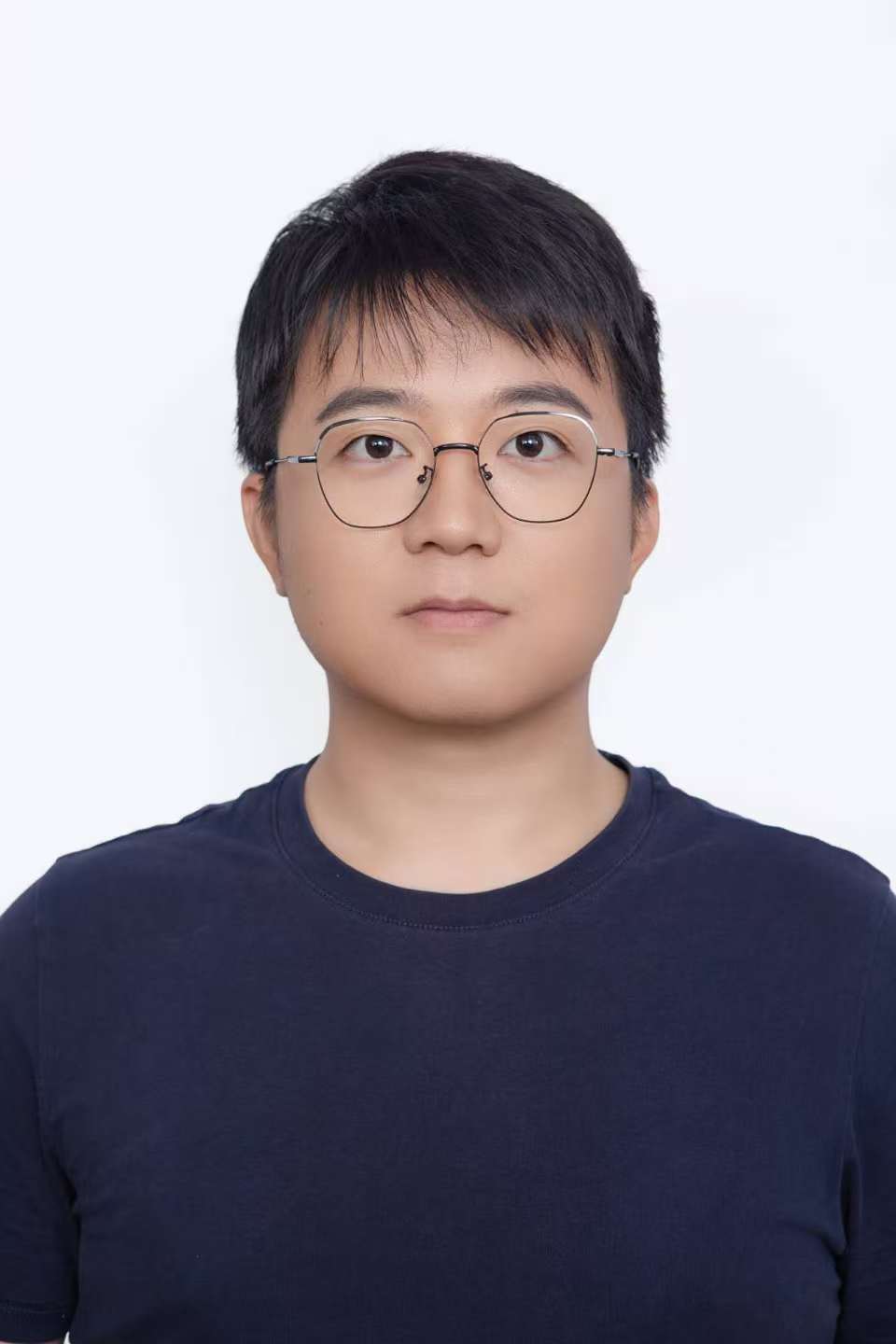}}]{Xinyan Liu} is currently an Associate Professor at Harbin Institute of Technology (HIT) and a Postdoctoral Fellow at City University of Hong Kong. He received his Ph.D. degree from the University of Chinese Academy of Sciences (UCAS) in 2025. His research interests lie at the intersection of video understanding and cross-media content understanding, with a focus on multimodal learning, video analysis, and intelligent content processing. He has published several academic papers in top-tier journals and conferences in her field. His current research explores the integration of computer vision and multimodal data analysis for advanced media intelligence applications.
\end{IEEEbiography}

\vspace{-400pt}

\begin{IEEEbiography}[{\includegraphics[width=1in,height=1.25in,clip,keepaspectratio]{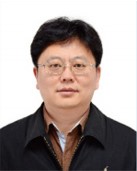}}]{Yongdong Zhang} (M’08–SM’13-F’24) received the Ph.D. degree in electronic engineering from Tianjin University, Tianjin, China, in 2002. He is currently a Professor with the School of Information Science and Technology, University of Science and Technology of China. His current research interests are in the fields of multimedia content analysis and understanding, multimedia content security, video encoding, and streaming media technology. He has authored over 200 refereed journal and conference papers, accumulating more than 29,000 citations on Google Scholar. He was a recipient of the best paper awards in PCM 2013, ICIMCS 2013, ICME 2010, the best student paper award in ACM Multimedia 2022 and the Best Paper Candidate in ICME 2011. He serves as an Editorial Board Member of the Multimedia Systems Journal and the IEEE TRANSACTIONS ON MULTIMEDIA. He is a fellow of the IEEE.
\end{IEEEbiography}

\end{document}